\documentclass{article}

\usepackage{arxiv}

\usepackage[utf8]{inputenc} 
\usepackage[T1]{fontenc}    
\usepackage{hyperref}       
\usepackage{url}            
\usepackage{booktabs}       
\usepackage{amsfonts} 
\usepackage{amsmath,amssymb,amsthm}
\usepackage{nicefrac}       
\usepackage{microtype}      
\usepackage{lipsum}
\usepackage{graphicx}
\usepackage{sidecap}
\sidecaptionvpos{figure}{c}
\graphicspath{ {./images/} }

\newcommand\ringring[1]{%
  {
   \mathop{\kern0pt #1}\limits^{
     \vbox to-1.85ex{
       \kern-2ex 
       \hbox to 0pt{\hss\normalfont\kern.1em \r{}\kern-.45em \r{}\hss}%
       \vss 
     }
   }
  }
}

\title{Field-mediated locomotor dynamics on highly deformable surfaces}

\author{
 Shengkai Li \\
  School of Physics\\ Georgia Institute of Technology\\ Atlanta, Georgia 30332\\
   \And
 Yasemin Ozkan Aydin \\
  Department of Electrical Engineering\\ University of Notre Dame\\ Notre Dame, Indiana 46556\\
  \And
 Charles Xiao \\
  Department of Mechanical Engineering\\ University of California Santa Barbara\\ Santa Barbara, California 93106 \\
  \And
 Gabriella Small \\
  Walker Department of Mechanical Engineering\\ The University of Texas at Austin\\ Austin, Texas 78712 \\
 \And
 Hussain N. Gynai \\
  School of Mathematics\\ Georgia Institute of Technology\\ Atlanta, Georgia 30332 \\
 \And
 Gongjie Li \\
  School of Physics\\ Georgia Institute of Technology\\ Atlanta, Georgia 30332\\
 \And
 Jennifer M. Rieser \\
  Department of Physics\\ Emory University\\ Atlanta, Georgia 30322 \\
 \And
 Pablo Laguna \\
  Center for Gravitational Physics\\ Department of Physics\\ University of Texas at Austin\\ Austin, Texas 78712\\
 \And
 Daniel I. Goldman \\
  School of Physics\\ Georgia Institute of Technology\\ Atlanta, Georgia 30332\\
  
}

\begin{document}
\maketitle
\begin{abstract}
Studies of active matter -- biological, robotic and computational systems consisting of individuals or ensembles of internally driven and damped locomotors -- are of much interest. While principles governing active systems in the bulk of fluids and flowable terrestrial environments (like sand) are well studied, another class of systems exists at interfaces. Such interfaces can be rigid and yielding solids (hard ground and sand), deformable fluids (air-water interface), and even elastic surfaces (e.g. leaf litter, spiderwebs). The latter cases are especially interesting as locomotors are strongly influenced by the geometry of the surface, which in itself is a curvature dominated dynamical entity. To gain insight into principles by which locomotors are influenced via a deformation field alone (and can influence other locomotors), we study robot locomotion on an elastic membrane, a model of active systems experiencing highly deformable interfaces. As a model active agent, we use a differential drive vehicle which drives straight on flat homogeneous surfaces, but reorients in response to environmental curvature. We study the curvature field mediated dynamics of a single vehicle interacting with a fixed deformation as well as multiple vehicles interacting with each other via local deformations. Single vehicle display precessing orbits in centrally deformed environments while multiple vehicles influence each other by local deformation fields. The active nature of the system facilitates a General Relativity inspired mathematical mapping from the vehicle dynamics to those of test particles in a fictitious ``spacetime'', allowing further understanding of the dynamics and how to control agent interactions. 
\end{abstract}


\section*{Introduction}
Study of systems composed of internally driven agents has long been the domain of biology \cite{Mccreery2016CollectiveAnts,Anderson2002SelfAssemblagesSocieties,Hemelrijk2012SchoolsSelf-organization} and robotics \cite{Brambilla2013SwarmPerspective,Schranz2020SwarmApplications,Chung2018ARobotics,Berlingereabd8668} 
but is coming into vogue in physics in the field of active matter \cite{RevModPhys.85.1143,aguilar2018collective,chvykov2021low}. Individual active agents display novel dynamics as well largely a function of their persistent dynamics \cite{bush2015pilot} and control \cite{hu2003hydrodynamics}. Active collectives display fascinating properties that non-active systems do not, for example, collectives can show non-reciprocal phase transitions \cite{fruchart2021non}, field drive \cite{wang2021emergent}, and can interact via physical (wakes in fluids) and social forces \cite{silverberg2013collective}. And in aero \cite{cavagna2014bird}, hydro \cite{parrish1997animal} and even terradynamic \cite{li2013terradynamics,aguilar2016review} interactions in the bulk are well explored and increasingly understood regarding locomotion such that we can build capable devices \cite{li2017fast,lau2020stunt} and create capable swarms \cite{Berman2011,Rubenstein795,Brambilla2013SwarmPerspective,Schranz2020SwarmApplications,Chung2018ARobotics}.

In contrast, interactions in which locomotors move on an environment which is highly but not necessarily permanently, deformed by the locomotor such that the environmental deformation field plays an important role in the locomotion and non-contact interactions are much less understood and relevant in many active systems across environments.  Such interactions are seen on the surface of fluids (e.g. water walking insects and robots \cite{hu2003hydrodynamics,hu2005meniscus} but also in terradynamic situations like in flowable environments of granular media \cite{li2013terradynamics,shrivastava2020material}, environments with gaps and holes \cite{qian2020obstacle, gart2018dynamic}, vertical posts \cite{rieser2019dynamics,li2015terradynamically}, and elastic surface like leaf litter \cite{spence2010insects}, running tracks~\cite{hayati2019effects,mcmahon1979influence}, and soft ground \cite{ferris1998running,ferris1999runners} and organisms running rapidly over muddy substrate (shear thickening substrates \cite{baumgarten2019general}). In addition, in such systems, field mediated interactions (like sensing through vibration and deformation of webs) have the feature that they can sense and influence other locomotors without direct contact \cite{Landolfa2004VibrationsIT,spider2014,Liang2018}. Better understanding of such terradynamic interactions, could lead to global control of multiple agents via local sensing of field interactions alone \cite{wang2021emergent,Berman2011,Dong2020,Elamvazhuthi_2019}.

To mitigate complexity of hydrodynamic and complex terradynamic surfaces, we chose as a model system an elastic membrane to study interaction of locomotor(s) on highly deformable environments. Specifically we study the locomotion and field interaction of vehicles on highly deformable elastic environments via the study of two cases: a single vehicle in the presence of a fixed obstacle and multiple (two) vehicles influencing each other. We find that the dynamics of even a single vehicle are interesting when influenced by a non-moving boundary via membrane curvature alone. Despite no sensing or control, the vehicle orbits, collides or escapes, analogous to how bodies orbit stars. We then show how multiple vehicles generically display a substrate mediated cohesion whose collisions timescale depends on vehicle mass, which is reminiscent of the cheerios effect \cite{vella2005cheerios}. Inspired by the response of the vehicle to curvature, we develop controllers which can help avoid such cohesion via local measurements of tilt thus indeed generating global control over local forcing. Inspired by the field mediated dynamics, we find that the active aspect of this system makes it possible to construct a mapping between the vehicle dynamics and the geodesics of a tunable spacetime with metric description. We use this theory to explain aspects of individual vehicle dynamics which are reminiscent of GR (e.g. precessing orbits) as well as demonstrate how our controller mitigates cohesion via measurements of its tilt and gets away from the deformation field.

\section*{Active agent on membrane: field mediated interaction from fixed object}\label{sec:setup}
\begin{SCfigure}[][ht!]
    \centering
    \includegraphics[width=0.6\textwidth]{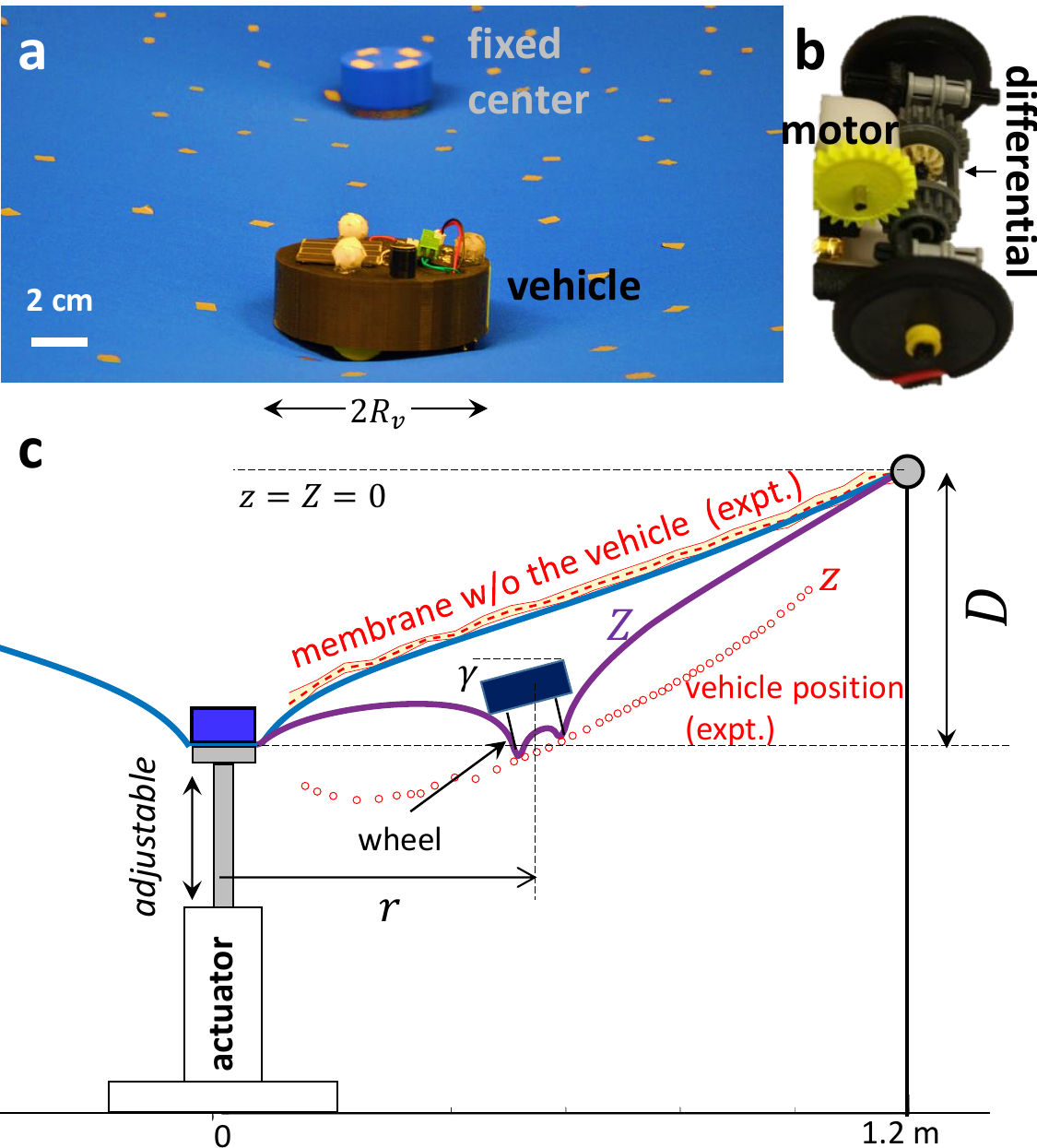}
    \caption{\textbf{Interaction induced by elastic substrate deformation.} (a) One vehicle transiting around a central depression. (b) Side view of the differential driven vehicle and side view of the drive mechanism; see also Fig. \ref{fig:vehicleMech}a for more details. (c) Cross section of the experimental set-up with a depression $D$ ($z$ axis of the membrane is linearly stretched for visual clarity). The red dotted line denotes the measurement of the membrane shape in the absence of the vehicle, the red open dots show the contact positions of the vehicle  with the membrane when it is placed at different radii. The vehicle's vertical position $z$ relative to the confining outer membrane ring is approximated by the average of the membrane height $Z$ around it: $z(\mathbf{r})\approx\langle Z(\mathbf{r}')\rangle_{|\mathbf{r}'-\mathbf{r}|=R_v}$. In the axi-symmetric case shown here, it can be further approximated by $z(r)\approx(Z(r-R_v)+Z(r+R_v))/2.$}
    \label{fig:membrane}
 \end{SCfigure}

We first study the dynamics of a differential drive vehicle self-propelling on a deformable curved surface. The vehicle (Fig.~\ref{fig:membrane}a,b) takes inspiration from many active matter experiments \cite{gart2011collective,zhang2010collective,lee2017self} and simulations \cite{vicsek1995novel,szabo2006phase,lee2017self} in that it moves straight in the absence of interaction. Further, its mechanics are key elements of modern wheeled vehicles, which are deployed in diverse terradynamic scenarios, from paved roads to Martian landscapes \cite{Hoogterp1999,Chung2001,WANG201163}. The vehicle has two rear wheels and one front spherical caster for stability. A critical feature of the vehicle is a {\em differential} ~\cite{uicker2011theory} which allows independent rotation of the wheels upon different load conditions by maintaining constant speed governed by motor rotation rate. If the load of the two wheels is equal, e.g. the vehicle is on level ground, both wheels turn at the same rate and the vehicle goes straight. If the load of one of the wheels increases (i.e. vehicle tilts) the corresponding wheel slows down and the opposite wheel speeds up, which results in turning motion around the slow wheel. While it seems we have used a particular robot to perform the study, we note that constant speed motion is a convenient starting point to study more general dynamics in such active systems.

Experiments were performed on a four-way stretchable spandex fabric (that stretches and recovers both width and lengthwise) affixed unstretched to a circular metal frame (see Materials and Methods) with a radius of $R=1.2$~m. In the first situation with a fixed center, a linear actuator attached to the center of the membrane warps the fabric from underneath to allow adjustable central depression of depth $D$ with a cap (radius $R_v=5$ cm) fastening the actuator to the fabric. A diagram of the experimental setup is given in Fig.~\ref{fig:membrane}c.  The membrane has a measured axi-symmetry such that the standard deviation of the membrane height at each radius is less than $5 \%$ of the central depression magnitude $D$ (see S8.2 of the SI).

Three aspects are important to understand the dynamics of the vehicle on the membrane. The first is that the vehicle dynamics are highly damped and inertia plays a minimal role: if the motor stops the vehicle rapidly comes to rest (within a second). That is, there is no ``rolling down hill''. The second aspect is that the differential in the vehicle allows it to turn dynamically according to the local curvature instead of simply following the spatial geodesics of the membrane, which leads to almost straight trajectories given the shallow depressions of the membrane (see S3 in the SI). The third important aspect is that, while the global shape of the membrane without the vehicle is important, due to the vehicle's mass, its local environment deviates from the ``bare'' shape of the membrane, introducing an additional local deformation of the membrane. This results in a vehicle tilting to an angle $\gamma$ (between the normal of the vehicle surface and $\hat{z}$) depending on the vehicle's radial position in membrane as depicted in Fig.~\ref{fig:membrane}c.

\begin{SCfigure}[][ht!]
    \centering
    \includegraphics[width=0.45\textwidth]{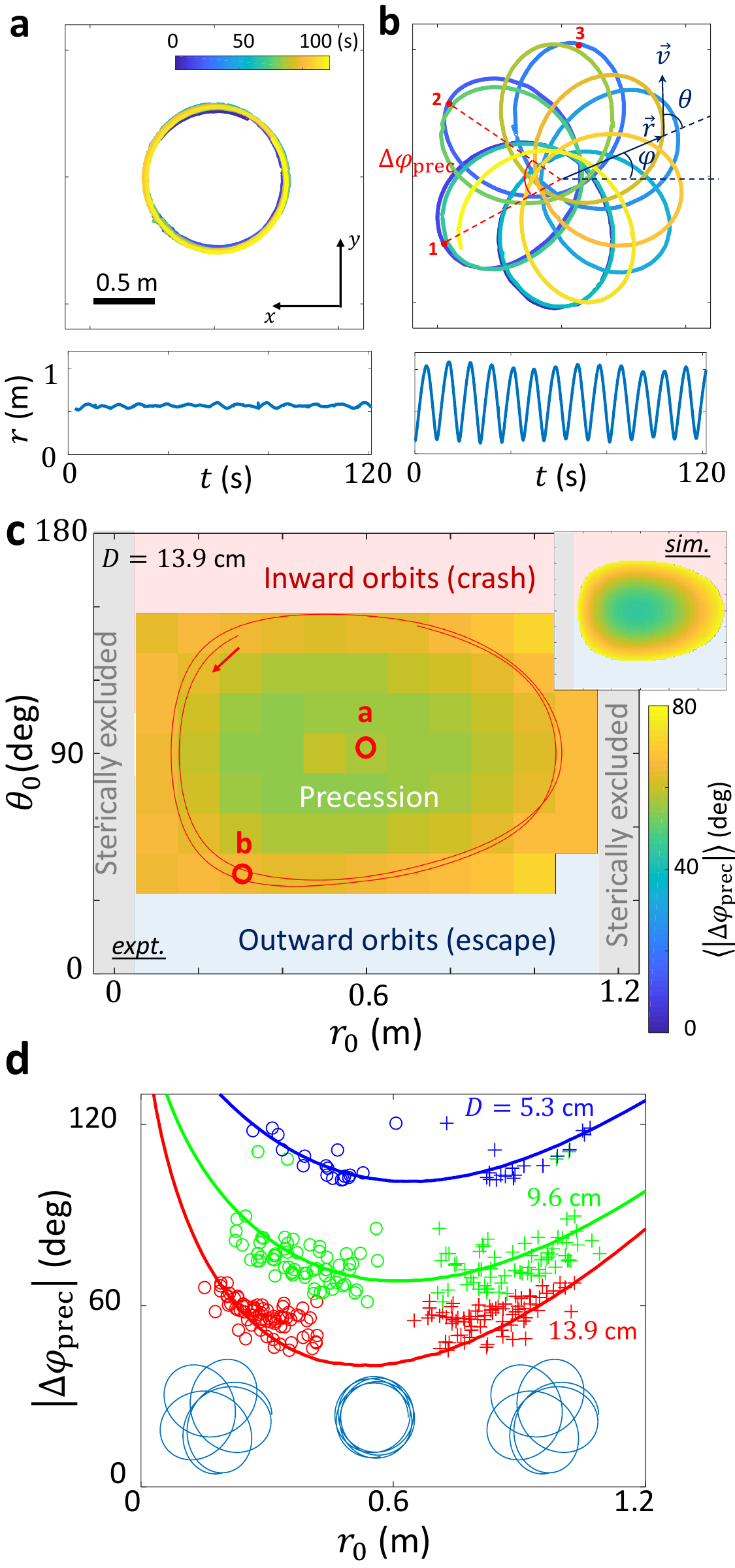}
    \caption{\textbf{Examples of bound vehicle trajectories.} An example (a) circular orbit (see Movie S1.mp4) and (b) eccentric orbit for the central depression $D = 13.9$~cm (see Movie S2.mp4). In (b) the angle $\theta$ denotes the heading angle and $\varphi$ denotes the azimuthal angle. The corresponding evolution of the radius over time are shown below. The eccentric orbit exhibits a precession of $|\Delta \varphi_{\text{prec}}| \approx \pi/3$ evaluated from consecutive apoapsis or periapsis (peaks or valleys on the $r-t$ plots). (c) Precession angle's dependence on initial condition. The initial condition of the circular orbit (a) is indicated by a red circle. Any points on the trajectory of (b) can be considered as an initial condition of it. Two orbits of (b) are shown in a red curve. The inset shows the prediction from theory; axes ranges are the same as those in main figure. (d) Precession angle $|\Delta\varphi_{\text{prec}}|$ as a function of the effective initial radius $r_0$ for $\theta_0=\pi/2$ and  central depressions $D=13.9$~cm (red), $9.6$~cm (green), and $5.3$~cm (blue). Experimental data are dots and solid lines are theoretical prediction using Eq.~(\ref{eq:du2}). The open dots show the $r_{-}$ and the pluses show the $r_{+}$. The insets below the curves show the trajectories at different radii and angular momenta for $D=9.6$ cm.}
    \label{fig:precession}
 \end{SCfigure}

For simplicity, we first study the dynamics of a single vehicle moving at constant speed on the membrane (set by constant motor rotation rate and enforced by the differential mechanism). Experiments were conducted by setting the initial radius $r$ (the distance between the center of the vehicle and the setup) and heading angle $\theta$ (the angle between the radial direction and the velocity of the vehicle, Fig.~\ref{fig:precession}b). The trajectories of the vehicle were recorded for 2 minutes by a high-speed motion capture system (Optitrack, 120 Hz) positioned above the membrane. Certain initial conditions (a particular radius $r_0=r_c \approx 0.6$m and heading $\theta_0 \approx 90^\circ$) developed immediate circular orbits (Fig.~\ref{fig:precession}a). However, similar to the orbiting droplets on the liquid surface curved by their weight \cite{gauthier2019capillary}, for a majority of ($r_0, \theta_0$), we observed trajectories consisting of {\em retrograde} precessing ellipse-like orbits (Fig.\ref{fig:precession}b) about the central depression, i.e. the maximum radius of the orbit does not return to the same azimuthal position but rather lags behind after an orbit. 

The precessing dynamics can persist for many orbits until the vehicle's orbit either slowly increases or decreases its eccentricity. In the former case, the vehicle ultimately collides with the central cap or escapes to the boundary. In the latter case, the precession decays into an approximately circular orbit with a critical $r_c$ radius depending on the central depression $D$. From analysis of the vehicle mechanism and dynamics (see S2 in SI), we attribute these behaviors to slight mechanical imperfections in the mass distribution in the vehicle, such as the deviation of the center of mass from the center-line, $\Delta B$. The eccentricity evolves over orbits with a factor $e^{-\epsilon\varphi/2}$ where $\epsilon=\frac{L_c \Delta B}{\frac{1}{2}R_v^2+L_c^2+\Delta B^2}$; the precessing dynamics can observed over longer timescales as the magnitude of imperfection decreases. Here $L_c$ is the distance between the wheel axle and the center of mass, and $R_v$ is the radius of the vehicle. Ideally, a perfect vehicle with $\Delta B = 0$ makes $e^{-\epsilon\varphi/2}$ remain at $1$ and the orbit stays in the steady-state forever. The half life $(2\log{2})/\epsilon$ characterizes how steady an orbit is; the sign of $\Delta B$ determines if the eccentricity will expand or decay.

To gain insight into the precessing dynamics, for bounded steady-state trajectories with half-lives of eccentricity longer than $5$ revolutions, we measured average precession $|\Delta \varphi_{\text{prec}}|$ as a function of initial conditions ($r_0, \theta_0$) by evaluating the change in angular location of consecutive apoapsides or periapsides (e.g. between periapsis 1 and 2 in Fig.~\ref{fig:precession}b). A map of this is shown in Fig.~\ref{fig:precession}c in the $r-\theta$ space. We choose the heading angle $\theta$ rather than the azimuthal angle $\varphi$ to reduce the redundant counting of the same trajectories shifted by just an azimuthal angle due to the axi-symmetry. We find the precession angle to be constant throughout the trajectory, therefore all the points sampled from a trajectory share a constant precession angle and each point's $(r,\theta)$ along this trajectory can be regarded as an effective initial condition in the trajectory $r$-$\theta$ space. Including these initial conditions, the map reveals that the precession is minimal when the vehicle is initiated at a particular radius $r_c$ ($\approx 0.6$ m when the central depression $D=13.9$ cm) and heading of $90^\circ$; $|\Delta \varphi_{\text{prec}}|$ increased as initial conditions deviated from this region. However, $r_0$ is restricted to the range $0.2$~m $ \le r_0 \le 1.1$~m to exclude the central cap in the membrane and to avoid starting the vehicle too close (less than $10$ cm) to the outer ring. Initial headings which pointed approximately towards or away from central depression did not achieve a stable orbit. That is, for $\theta_0<30^\circ$ the vehicle collided with the outer boundary, and for $\theta_0>150^\circ$ the vehicle crashed into the central cap.

\section*{Minimal model for vehicle dynamics}
\begin{SCfigure}[][ht!]
    \centering
    \includegraphics[width=0.35\textwidth]{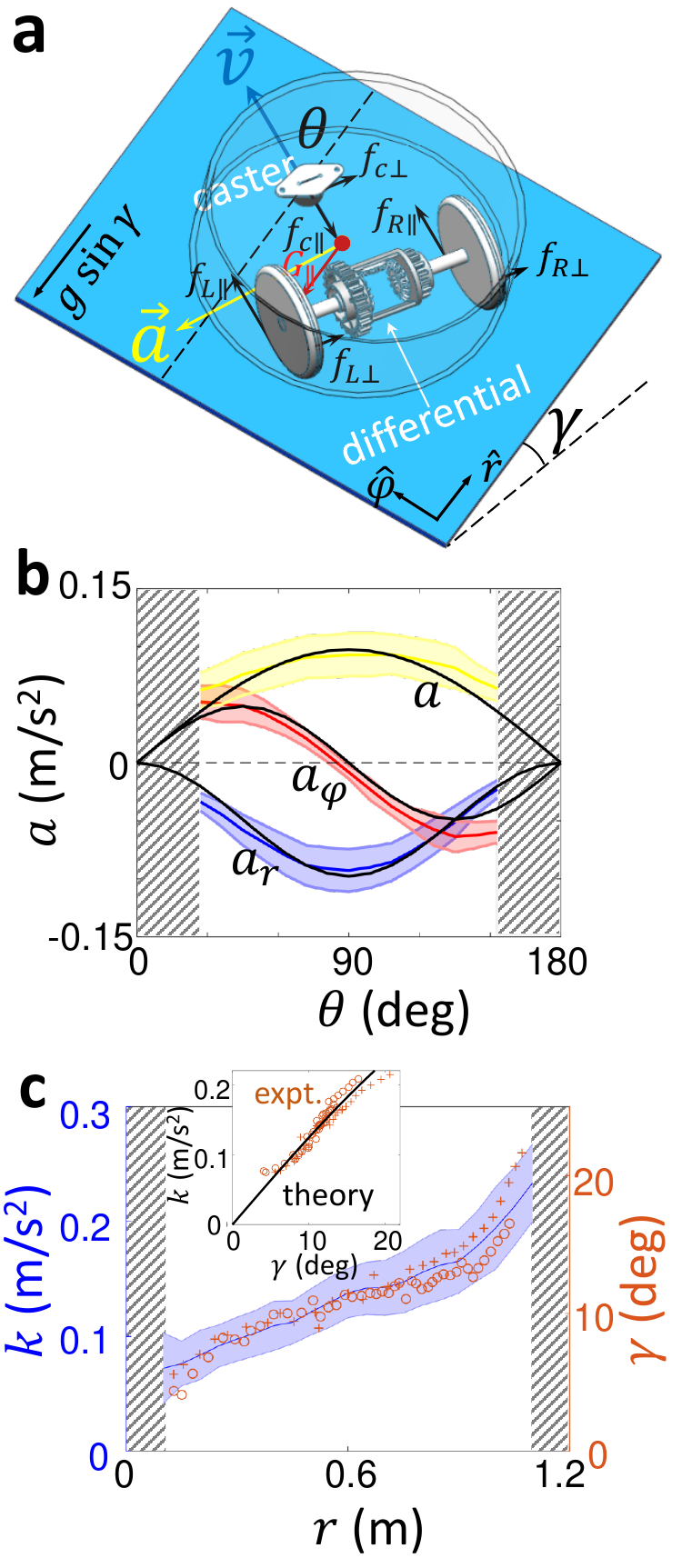}
    \caption{\textbf{Vehicle dynamics.} (a) Schematic of a vehicle moving on a piece of membrane and its force diagram. The dashed line on the incline shows one of the membrane radii. The frictions on the wheels and the caster are shown in black and the component of Earth gravity along the slope is shown in red. (b) Magnitude of the acceleration $a$ (yellow), and its components $a_r$ (blue) and $a_\varphi$ (red) as a function of the heading angle $\theta$ evaluated at $r = 0.3$~m (central depression $D=9.6$ cm). Black lines correspond to $a = k(0.3\text{m})\cdot\sin{\theta}$, $a_r = -a\sin{\theta}$, and $a_{\varphi}= a \cos{\theta}$. The gray shaded regions indicate extreme headings that do not have steady trajectories. (c) The acceleration function $k$ and vehicle tilt $\gamma$ as a function of the radius $r$ for $\theta = 90^\circ$ with the solid blue line and shading denoting the mean and standard deviation of $k$ obtained from the experiments. The red markers show the vehicle tilt $\gamma$ measured from the experiment on two different azimuths separated by $90^\circ$ with open circles and pluses respectively. Inset shows the relation between $k$ and $\gamma$ using the $k$ data from the main figure and the theoretical curve $k=0.074~g \sin{\gamma}\cos{\gamma}$.}
    \label{fig:vehicleMech}
 \end{SCfigure}

To gain insight into how vehicle orbital dynamics emerge solely due to interaction with the curvature field generated by both the central depression and the vehicle's local depression field (Fig.~\ref{fig:membrane}a), rather than solving a coupled membrane-vehicle interaction system of equations, we instead construct a minimal model which gives physical insight into how a such deformation fields influence the vehicle's dynamics. Since the vehicle moves at a constant speed this requires that the acceleration is perpendicular to the velocity $(v^r,v^{\varphi})$ such that
\begin{eqnarray}
 a^{\varphi} &=&\ddot{\varphi} + \frac{2\,\dot r \,\dot\varphi}{r} = \frac{a\,v^r}{r\,v} =  \frac{a}{r}\cos{\theta}\label{eq:reformulate1} \\
a^r &=&      \ddot{r} - r \,\dot{\varphi}^2 = -\frac{a\,r\,v^\varphi}{v} = -a\sin{\theta}\label{eq:reformulate2}\,.
\end{eqnarray}
with dots denoting differentiation with respect to $t$. Our experiments reveal that, to a good approximation, the vehicle's acceleration is given by $a= k\,\sin{\theta}$ where $k$ is a function of $r$ only (Fig.~\ref{fig:vehicleMech}c). We want to point out the form of $k\sin{\theta}$ can be regarded as the first order expansion of any $a(r,\theta)$ for any vehicles. Having $a \propto \sin{\theta}$ implies that, in this axi-symmetric case, the magnitude of the acceleration is proportional to that of the cross product between the velocity and the gradient of the terrain since the gradient of the terrain is aligned with the radial direction. We will later show that $a \propto |\hat{v} \times \nabla z |$, where $z$ denoting the vertical position of the vehicle from the frame holding the spandex sheet, also holds for surfaces with arbitrary shapes. 

We treat the vehicle with tilt angle $\gamma$ from its level orientation as driving on a local incline with slope $\gamma$. From a theoretical analysis of how the constant-speed differentially driven vehicle pivots on a slope  (Fig.~\ref{fig:vehicleMech}b, see S1 of SI for derivations), we found that $k = C\,g\, \sin{\gamma}\,\cos{\gamma}\approx C\,g\, |\nabla z|$ with $g$ Earth's gravity. The prefactor $C$ is a mechanical constant related to the structure of the vehicle as $C=L_c^2/(L_c^2+\frac{1}{2} R_v^2)$ where $L_c\approx 1~$ cm is the distance between the wheel axle and the center of mass (see Fig.S1), and $R_v=5$~cm is the radius of the vehicle. The theoretical value for $C$ from the model is approximately $0.074$, while the experimental fit (Fig.~\ref{fig:vehicleMech}c inset) gives a value of $0.073\pm 0.001$ (see S1 of SI). 

The  model as described by Eqs.~(\ref{eq:reformulate1}) and (\ref{eq:reformulate2}) yields good agreement with experiments over a range of $v=0.20-0.32$ m/s. The essential ingredient of the model is that the differential mechanism ensures torque balance on both wheels. In addition, the rolling friction on the caster is negligible compared to other contact forces (see Fig.~\ref{fig:vehicleMech}a for force diagram). The model indicates $k=a/\sin{\theta}$ should be the same for any $\theta$ for a balanced vehicle. The experimentally measured result shows a slight dependence on heading angle $\theta$ (Fig.S1 in SI) that can be understood as weight imbalance, characterized by $\Delta B$. Introduction of this bias into the analysis returns a correction in the form of $a_{bias}/\sin{\theta}=k\cdot (\Delta B/L_c)\cot{\theta}$. It vanishes when $\theta=\pi/2$ or $\Delta B=0$ (perfectly balanced vehicle). Integration of this yields precession dynamics that quantitatively matches with the experiments for all different depressions (Fig. \ref{fig:precession}c,d). 

An important aspect of the dynamics which is revealed by the model is that unexpectedly the vehicle does not follow spatial geodesics of the membrane (as in the museum demos of GR \cite{possel2018relatively}) given by curves with
\begin{eqnarray}
ds^2=\Psi^2 dr^2+r^2 d\varphi^2 \label{eq:spaGeod}
\end{eqnarray}
where $ \Psi^2\equiv 1+z'^2$ and $z'\equiv \partial z/\partial r$ is the gradient of the vehicle's height $z$. These spatial-only geodesics are nearly straight lines in our setup (see Fig.S3 in SI). We will later show that the essential ingredients that generate this difference are the deformation of membrane by the vehicle, and the active nature of the system shown in the vehicle's ability to change the direction of motion as a consequence of the local tilt of the vehicle. These two aspects are reflected respectively in the acceleration strength $k$ governed by membrane deformation and $\theta$, the heading angle of the vehicle.

\section*{Understanding the dynamic features through spacetime mapping}
\label{sec:spacetime}

Given that the vehicle's precessing orbits emerge solely from curvature field mediated interaction, and that such dynamics are reminiscent of orbiting  bodies in curved spacetimes, we wondered if the framework of general relativity could give further insight into the vehicle's dynamics. And inspired by the observation that the orbits we observe do not follow geodesics of the membrane, it is interesting to ask whether there exists a map that recasts the dynamics of the active vehicle in physical space into that of the geodesic motion of a so-called ``test particle'' in an effective space-time. Recall that in GR, test particles follow geodesics of a spacetime governed by a so-called ``metric'' which is a rule specifying distances between close points in spacetime. Physics has a history of explaining phenomena by mapping to a different space which can help give insight into dynamics in complex situations -- e.g.differential equations that could be easily solved in the Fourier space while direct solution in the real space is hard \cite{stein2011fourier}, and complicated motion planning which can be simply interpreted with the curvature in a shape space \cite{hatton2013geometric}.

In principle, the vehicle dynamics we wish to map could be described by a diversity of metrics. But for simplicity, and to make the analogy with GR, given the axi-symmetry of the system, we propose a metric of the form
\begin{equation}
  ds^2 = -\alpha^2 dt^2 +\Phi^2(\Psi^2 dr^2 + r^2 d\varphi^2)
  \label{eq:metrics}
\end{equation}
with $\alpha = \alpha(r)$, $\Phi = \Phi(r), \Psi^2=1+z'^2$. With the metric (\ref{eq:metrics}), the geodesic equations take the following form: 
\begin{eqnarray}
 \ddot{\varphi} + \frac{2\dot r \dot\varphi}{r} &=&  \left[\frac{(\alpha^2)'}{\alpha^2} - \frac{(\Phi^2)'}{\Phi^2}\right]\dot r\,\dot\varphi\label{eq:Geo1} \\
\ddot{r} - \frac{r\dot{\varphi}^2}{\Psi^2}+\frac{\Psi'}{\Psi}\dot{r}^2&=&  \left[\frac{(\alpha^2)'}{\alpha^2} - \frac{(\Phi^2)'}{\Phi^2} \right]\dot r^2\nonumber\\
&+& \frac{ (\Phi^2)'v^2-(\alpha^2)'}{2\,\Phi^2\Psi^2}\,. \label{eq:Geo2}
\end{eqnarray}
where primes denoting differentiation with respect to $r$.

Notice that the left hand side of Eqs.~(\ref{eq:Geo1}) and (\ref{eq:Geo2}) are the components of the acceleration, $a^\varphi$ and $a^r$ respectively, in  Eqs. (\ref{eq:reformulate1}) and (\ref{eq:reformulate2}). Thus, comparing the right hand side of these equations yields the following relationships between the metric functions $\alpha$ and $\Phi$ in terms of the speed of the vehicle and $k$:
\begin{eqnarray}
\alpha^2 &=&  E^2(1-v^2 e^{-K/v^2})\label{eq:alpha1}\\
\Phi^2 &=& E^2 e^{-K/v^2} (1-v^2 e^{-K/v^2})\label{eq:Phi1}
\label{eq:metric}
\end{eqnarray}
where $K(r)\equiv \int_0^r k(s)\Psi(s)ds$ and the constants of integration were chosen such that when $k=0$, the metric is flat (see S5 of SI). The quantity $E$ is a constant of motion (energy) associated with the fact that the metric is time-independent. The other constant of motion is $L$ (angular momentum) associated with the metric $\varphi$-symmetry.

Thus our formulation indeed reveals that the vehicle does not simply follow spatial geodesics of the membrane but instead follows geodesics in an emergent spacetime generated by the global curvature, the local curvature, the active dynamics and the differential mechanism.  The resultant dynamics can now be understood as test particle in a new spacetime where the active feature generates an effective spacetime not that of splittable space-times ( i.e. $g_{tt}=\alpha^2$ is constant) in the language of the work by Price~\cite{price2016spatial} in which the effects of curvature are restricted to space \cite{batlle2019exploring}. The essential contribution from active drive is the persistent response to the local curvature, here particularly enabled by the controlled constant speed unseen in passive systems. In fact, when the response of the turning to the local slope vanishes at the limit $v\rightarrow \infty$ such that $\alpha^2 = \Phi^2 = E^2(1-v^2)$, the metric Eq.\ref{eq:metrics} with components Eq.\ref{eq:alpha1},\ref{eq:Phi1} reduces to a splittable (and flat) spacetime Eq.\ref{eq:spaGeod}. On the other hand, when $v$ is finite and controllable, the active locomotion provides more flexibility and programmability in fabricating the desired spacetime depicted by GR than the passive agents studied in the previous works such as the dissipative marbles \cite{middleton2014circular,middleton2016elliptical} rolling on a membrane.

As with the Schwarzschild solution, we can use the normalization of space-time velocity to investigate the type of orbits. In terms of the constants of motion $E$ and $L$, this condition reads $1 = \frac{\Phi^2}{\alpha^2}\Psi^2\dot r^2 + \frac{1}{r^2}\frac{\alpha^2}{\Phi^2}\frac{L^2}{E^2}+\frac{\alpha^2}{E^2} \label{eq:r3}$. This expression can be rewritten in the following suggestive form:
${\cal E} = \frac{1}{2}m\,\dot r^2 + V$,
with ${\cal E} = 1/2$, $m =\Phi^2\Psi^2/\alpha^2$ the effective mass, and $V =[ \alpha^2\,\ell^2/(\Phi^2 r^2)+\alpha^2/E^2]/2$ the effective potential, where we have defined $\ell \equiv L/E$. With the help of Eqs.~(\ref{eq:alpha1}) and (\ref{eq:Phi1}), this effective potential reads
\begin{eqnarray}
V &=& \frac{1}{2}\left(\frac{\ell^2}{r^2} e^{K/v^2} +1-v^2e^{-K/v^2}\right).
\label{eq:effPot}
\end{eqnarray}

\begin{SCfigure}[][ht!]
    \centering
    \includegraphics[width=0.45\textwidth]{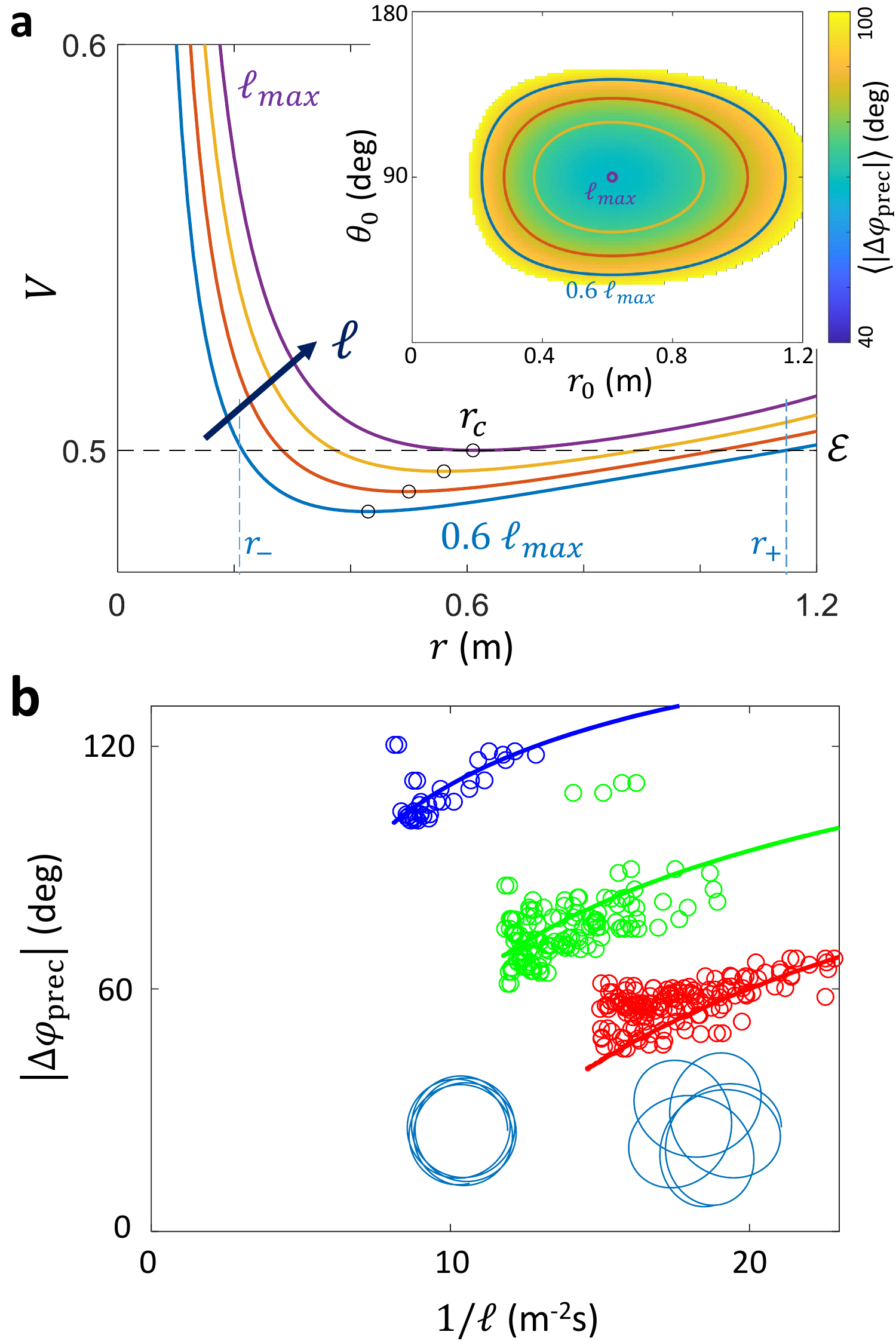}
    \caption{\textbf{The space-time derived effective potential governing the vehicle dynamics.} (a) $V$ is shown for different values of $\ell$ with $D=9.6~$cm. Black dots denote the minimum point of a given potential curve, and $r_c=v^2/r_c$ labels the case of a circular orbits when ${\cal E} = V_c$. The corresponding trajectories in the $r$-$\theta$ space are shown in the inset. (b) Precession angle $|\Delta\varphi_{\text{prec}}|$ as a function of the inverse of the relative angular momentum $\ell$ for the same cases. The insets below the curves show the trajectories at different radii and angular momenta for $D=9.6$ cm.}
    \label{fig:potential}
 \end{SCfigure}

Note that the energy and angular momentum enter through the ratio $\ell = L/E$, which can be calculated from the initial conditions since $\ell = \Phi^2 r^2\dot\varphi/\alpha^2$. Fig.~\ref{fig:potential}a shows examples of the potential $V$ for different values of $\ell$ with $\ell_{max}=v~r_c \exp{(-K(r_c)/v^2)}$ (see S6 in SI). The dashed line at $1/2$ denotes ${\cal E}$, and the turning points are given by  the solution to $r_{\pm} = \ell\, e^{K_{\pm}/v^2}/v$, where we use the subscript $\pm$ to denote a quantity evaluated at the turning points. Circular orbits occur when the minimum of the potential matches ${\cal E}$. The minimum is found from $V'=0$ and is located at $r_c = v^2/k_c$.\\

With the effective potential discovered from the mapping scheme, we can now explain the dependence of orbital precession on initial conditions and system parameters. To begin, we introduce the definitions of $E$ and $L$ to eliminate $\dot r$ in ${\cal E} = \frac{1}{2}m\,\dot r^2 + V$ in favor of $dr/d\varphi$. This results in
\begin{eqnarray}
\frac{\ell^2}{r^2}\left[\frac{1}{r^2}\left(\frac{dr}{d\varphi}\right)^2 +1\right] = v^2\,e^{-2K/v^2}\,.\label{eq:du2}
\end{eqnarray}

Next, we apply the change of variable $u = \ell/r$, and differentiate with respect to $\varphi$ and get
\begin{eqnarray}
\frac{d^2u}{d\varphi^2} + u = \frac{k\,\ell}{u^2}e^{-2K/v^2}\,.\label{eq:ddu}
\end{eqnarray}

As noted above, for circular orbits $r_c = v^2/k_c$, or equivalently $u_c = k_c\,\ell/v^2$ where $k_c \equiv k(r_c)$.
Perturbing Eq.~(\ref{eq:ddu}) about a circular orbit, i.e. $u = u_c + \delta u$, we get
\begin{eqnarray}
\frac{d^2\delta u}{d\varphi^2} + \left(1+\frac{k_c'}{k_c}r_c \right)\delta u = 0\,.
\end{eqnarray}
Thus, $\delta u \propto \cos{(\omega\,\varphi)}$ with $\omega^2 \equiv 1+r_c\,k'_c/k_c$, and the perturbative solution to Eq.~(\ref{eq:ddu}) is then given by
$u = u_c[1+e\,\cos{(\omega\,\varphi)}]$ where $e$ is the eccentricity of the orbit. Notice from this solution that one radial cycle takes place over a $2\pi/\omega$ angular cycle. Therefore, the precession angle is given by $\Delta\varphi_{\text{prec}} =2\pi/\omega_c-2\,\pi \approx -\pi\,r_c\,k'_c/k_c$.
Since $k_c > 0$, the sign of $\Delta\varphi_{\text{prec}}$, namely the direction of the precession, is given by the sign of $k_c'$ . If $k_c' > 0$, we have $\Delta\varphi_{\text{prec}} < 0$, retrograde precession, with prograde precession if $k_c' < 0$. From Fig.~\ref{fig:vehicleMech}c, we have that $k_c' > 0$, which explains the observed retrograde precession. Further, the dependence of $\Delta\varphi_{\text{prec}}$ with $r_c$ is consistent with our observation that the magnitude of the apsidal precession ($\Delta\varphi_{\text{prec}}<0$) decreases as the radius of the orbits approaches the radius of the circular orbit $r_c$.

We now reexamine the dependence of precession angle $\Delta\varphi_{\text{prec}}$ on initial conditions (Fig.~\ref{fig:precession}c) in the mapping framework. We now can see that contours of constant color correspond to trajectories with the same angular momentum $\ell$. And notably, the precession angle decreases as the orbits become more circular, with $\Delta\varphi_{\text{prec}}=-\pi\,r_c\,k'_c/k_c$ for the circular orbit. Fig.~\ref{fig:potential}b shows $\Delta\varphi_{\text{prec}}$ as a function of $r_0$ for initial heading angle $\theta_0 = 90^\circ$ with both the experimental data and the solution to Eq.~(\ref{eq:du2}). The minimum precession angle occurs for circular orbits. Again motivated by the Schwarzschild solution, for which $\Delta\varphi_{\text{prec}} = 6\pi G^2M/(c^2l)$ where $l\equiv\hat a(1-e^2)$ is the latus rectum, we evaluate the semi major-axis $\hat a$ and the eccentricity $e$ using the minimum and maximum radii: $\hat a=(r_{\text{max}}+r_{\text{min}})/2,~e=(r_{\text{max}}-r_{\text{min}})/(r_{\text{max}}+r_{\text{min}})$. Fig.~\ref{fig:potential}b shows $\Delta\varphi_{\text{prec}}$ as function of the inverse of the angular momentum $1/\ell\propto 1/\sqrt{l}$. While the trend is qualitatively similar to the Schwarzschild's solution connecting precession and eccentricity, in our metric, precession is never small and is not linear.

As a consequence of $k' >0$, our system generates retrograde orbits such that the vehicle's precession is opposite to that of GR in common situations. With our mapping, it is straightforward to understand how to obtain more GR-like prograde precession (like that of the Mercury\cite{Clifton2005,clemence1947relativity})): we must change the sign of the slope of $k$ so that $k'<0$ over a significant range of the vehicle trajectory. Because $k$ is connected to the tilting angle $\gamma$, we can achieve the desired change by increasing the tension of the membrane or decreasing the mass of the vehicle to enable the vehicle to more closely track the imposed membrane shape.

\begin{SCfigure}[][ht!]
    \centering
    \includegraphics[width=0.45\textwidth]{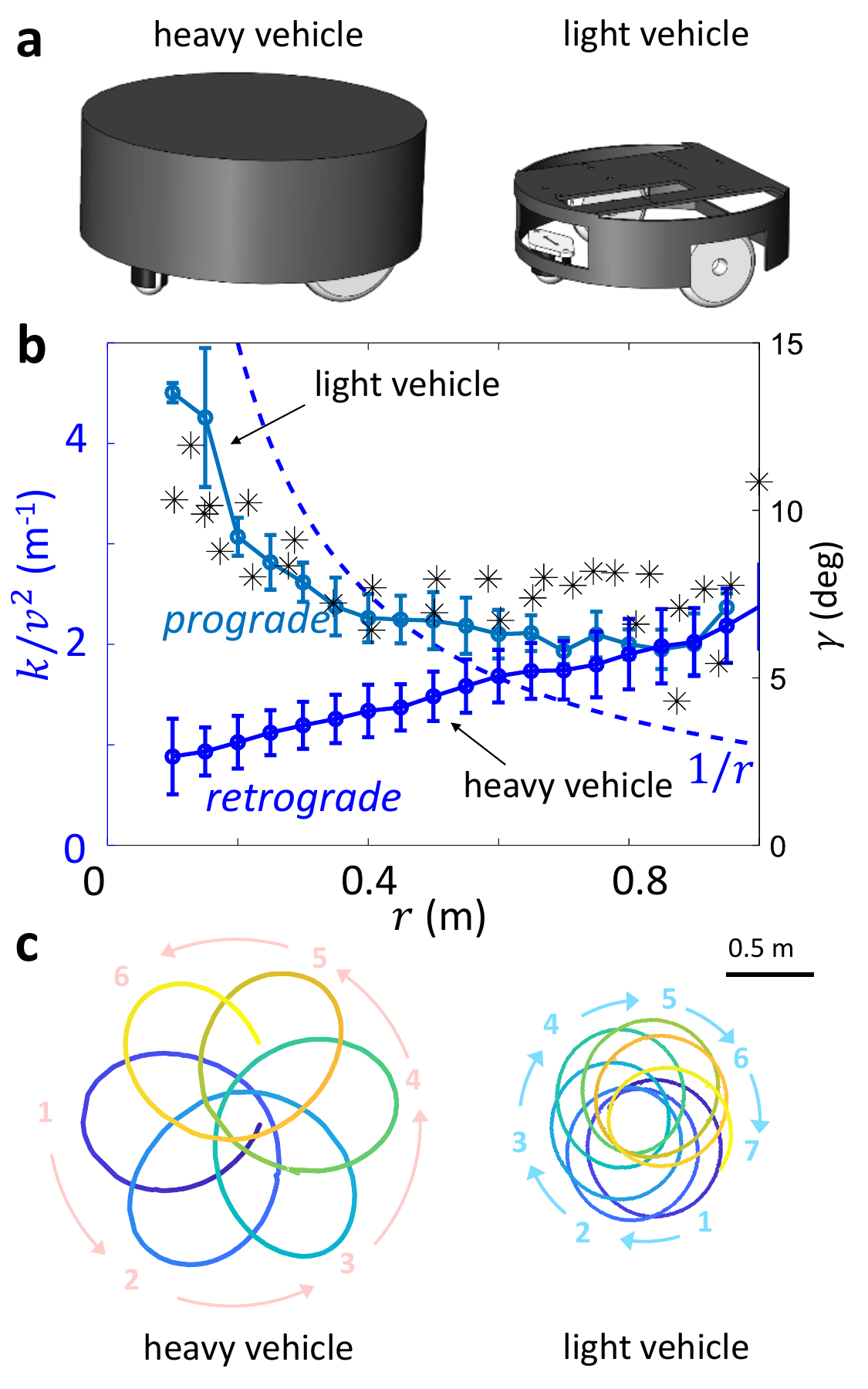}
    \caption{\textbf{Tuning the spacetime to generate prograde precession.} (a) The heavy ($m\approx 160$ g) vehicle and the light vehicle ($m\approx 45$ g). (b) Different $k(r)$ functions for prograde and retrograde precession. The light vehicle has a negative $k'$ at $r_c$ (the intersection of $k(r)/v^2$ and $1/r$) while the heavy vehicle has a positive $k'$ at such an intersection. The decreasing $k$ has the same trend as the tilt angle $\gamma(r)$ (black stars). (c) Clockwise trajectories with retrograde (left) and prograde (right) precessions. Perihelia are marked in order. The prograde precession is made by a lightweight vehicle on the membrane with $D=17$~cm central depression, for initial conditions $r_0=20$~cm$, \theta_0=90^\circ$. The periapsides numbered in blue show a clockwise order while the orbit is precessing in the same direction (For video, see SI3.mp4). The magnitude of the precession for this trial is $\Delta\varphi_{\text{prec}}=51^\circ$.}
    \label{fig:prograde}
 \end{SCfigure}

We chose to change the mass of the vehicle and constructed a smaller, lighter vehicle with mass $45$~g (Fig.~\ref{fig:prograde}a), approximately one quarter that of the original vehicle in Fig. \ref{fig:membrane}, a radius of $4$~cm, and a speed $v=0.11$ m/s. The vehicle produced trajectories (Fig.~\ref{fig:prograde}c) demonstrating prograde precession over all sampled initial conditions (65 total experiments). For a particular initial condition ($r_0=69$~cm$, \theta_0=90^\circ$), four out of five trials produced precessing orbits with significant eccentricity; here $\Delta\varphi_{\text{prec}}=+22^\circ \pm 16^\circ$. The theoretical prediction -- with $k(r)$ (Fig.~\ref{fig:prograde}b) deduced from such trajectories --  was $\Delta\varphi_{\text{prec}}=+33^\circ \pm 7^\circ$, within the experimental range. For a given initial condition, the lightweight vehicle showed greater trajectory variability than that of the heavier vehicle. We posit such variability is related to the slight membrane anisotropy, which makes the dynamics of the lightweight vehicle sensitive to initial conditions. Here the change of precession sign with the vehicle's mass demonstrates how the matter reciprocally tells the local spacetime how to curve and influences its global dynamical properties. We note that further modifications of the vehicle and the membrane could allow more analogous features comparing with the Schwarzschild metric, and this will enable studies in the strong field limit or on the modified GR theories using terradynamic experiments.

\section*{Development of a theory for reciprocal field mediated interaction dynamics}
\label{sec:SMI}

Thus far we have studied the interactions between a single vehicle and a central depression (which generates a time-independent imposed background) and have shown how we can understand the field-mediated orbital dynamics by mapping them to a ``spacetime'' using techniques from study of relativistically orbiting bodies. There are situations in which the deformation field experienced by a vehicle could be time-dependent; we have observed that a robot can be ``guided'' without contact via local deformation of the membrane alone (see SI4.mp4). Further, in the case of swarms of vehicles moving on a curvature field, this sets up interesting dynamics such that (in the case of two robots for example) each robot carries its own depression field and affects another other robot via this field alone (which could then affect the initial robot. Such a framework would embody Wheeler's succinct encapsulation of the reciprocal dynamics inherent in Einstein's view of gravity~\cite{wheeler1991geons, Ruffini1971,Misner1973}:  matter tells spacetime how to curve and spacetime tells matter how to move .

Therefore we next sought to develop a theory for the interaction of two agents via fields alone. The first element of the theory requires that we develop an equation of motion for the dynamics of a single vehicle experiencing an imposed deformation field that is not necessarily at the center of the membrane (the spacetime telling matter how to move component of the Wheeler encapsulation of the mechanics of GR). Thus we need to first generalize the equations of motion (Eq.\ref{eq:reformulate1},\ref{eq:reformulate2}) to that a vehicle on an arbitrary terrain. The axi-symmetric model can be generalized for arbitrary substrate by noticing that $a=k\,\sin{\theta}$ where $\theta$ is the angle between the velocity and the gradient of the slope and $k$ is the magnitude of the gradient timed by a mechanical constant. In the symmetric case, the gradient is always along the radial direction so that only the magnitude of the gradient $k=C\,g\,\sin{\gamma}\cos{\gamma}\approx C\,g\, |\nabla z|$ is needed. In the general case, noticing the $\sin{\theta}$ is the cross product of the unit vectors of the arbitrary terrain gradient $\mathbf{d}=-\nabla z$ and the vehicle velocity, the generalized equation of motion is
\begin{eqnarray}
\ddot{x}&=&C\,g\, \dot{y}\,(d_x \dot{y}-d_y \dot{x})/v^2 \label{eq:eomGen1}\\
\ddot{y}&=&-C\,g\, \dot{x}\,(d_x \dot{y}-d_y \dot{x})/v^2\label{eq:eomGen2}\,,
\end{eqnarray}
Conceptually, this is our ``$F=ma$'' with $\mathbf{d}$  playing the role of  ``$F$'' (recall $d_i = - \nabla_i z$ with $i = x,y$).

To complete the field mediated interaction picture, since a moving vehicle presents to another vehicle a time dependent deformation field, we require an equation to describe how a vehicle (the ``matter'') deforms the membrane curvature.

To characterize how the membrane responds to local perturbations, we use the wave equation, the simplest equation for a membrane assuming linear elasticity:

\begin{equation}
\ddot Z - v^2_m \Delta Z = -P\,.\label{eq:wave}
\end{equation}
where $v_m$ is the speed of propagation of disturbances in the membrane and $P = P_0\, (1+\tilde{P})$ with $P_0(>0)$ the force load from the membrane and $\tilde{P}$ the additional load from the vehicles, which is the area density of the vehicles normalized by the area density of the membrane. Since $P_0$ is the stationary force load when the membrane is only deformed by its weight, the time dependence in the source in Eq.~(\ref{eq:wave}) arises from $\tilde{P}$ due to the moving vehicles. Experiments examining the membrane elasticity have found that the shape of a free stationary membrane where $\tilde{P}=0$ and $\ddot{Z}=0$ follows the Poisson equation reasonably well (S8 in SI).

The speed of propagation for the membrane in our experiment is $v_m \approx 400$ cm/s, which is significantly larger than the typical speed of our vehicles ($v \approx 20$ cm/s). Therefore, we neglect time derivatives in Eq.~(\ref{eq:wave}) and solve instead the Poisson equation $\Delta Z = P/v_m^2$.

\begin{SCfigure}[][ht!]
    \centering
    \includegraphics[width=0.45\textwidth]{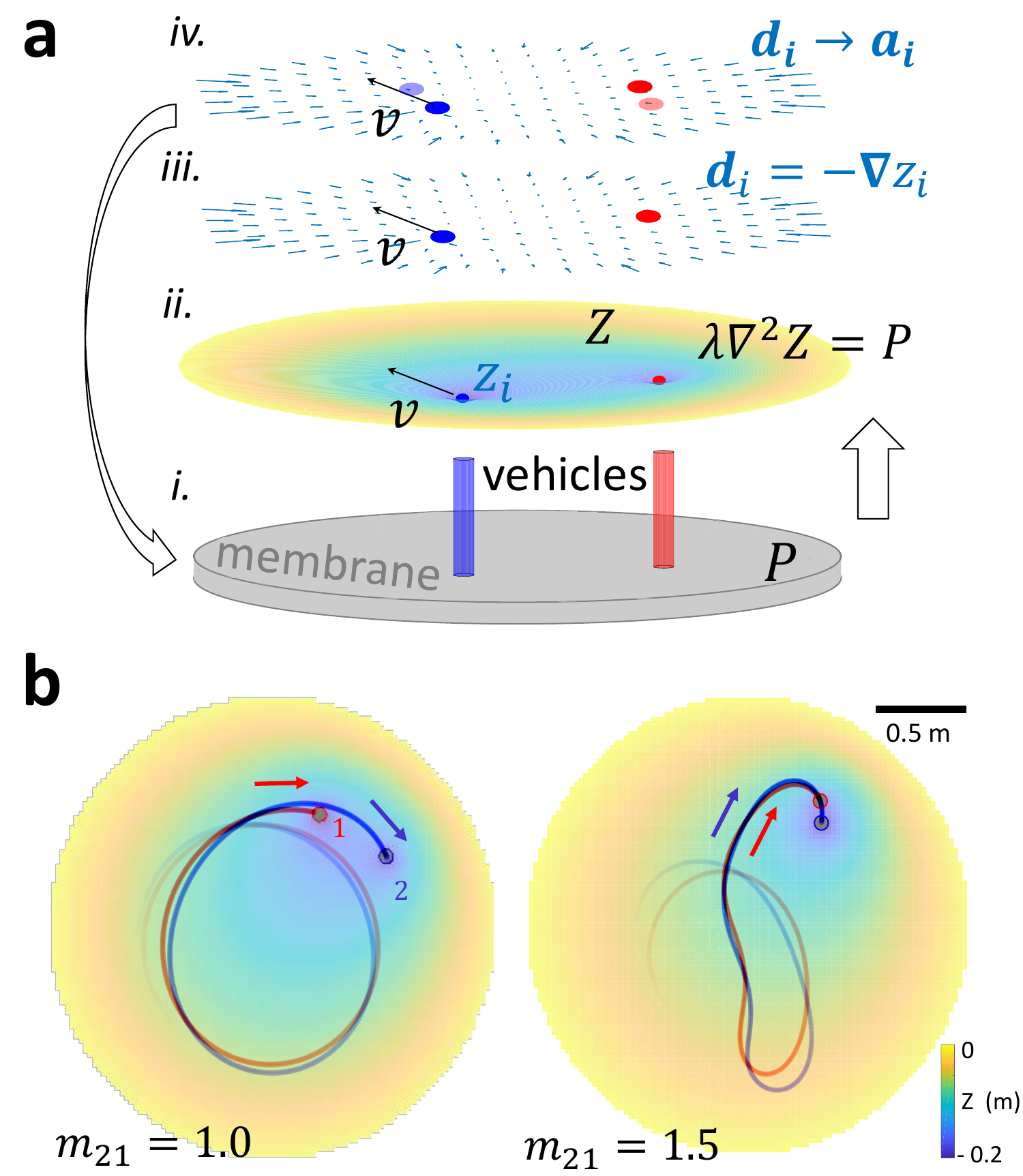}
    \caption{\textbf{Reciprocal interaction between the vehicle dynamics and curvature field.} (a) A sketch of the simulation procedure: (i) First, the shape of the membrane is solved from the Poisson equation with the load indicated in the bottom: gray disk for the membrane and two colored posts for the two vehicles in this example. (ii) Then, the height profile of the vehicle is evaluated at its position. (iii) Afterwards, the terrain gradient $\mathbf{d}$ is evaluated from the height profile of the vehicle. (iv) Finally, the acceleration determined by $\mathbf{d}$ using Eq.\ref{eq:eomGen1},\ref{eq:eomGen2} is integrated to update the new positions of the vehicles and the computation goes back to the first step again. (b) Theory predicts larger leader mass ratio ($m_{21}\equiv m_2/m_1 = 1.5$) fosters a merger better than a smaller one ($m_{21}=1.0$).}
    \label{fig:coupling}
 \end{SCfigure}

Therefore, the evolution of the system proceeds as follows (Fig.~\ref{fig:coupling}a): given the location of the vehicles, one first constructs the source $P$ and solves $\Delta Z = P/v_m^2$ to obtain the membrane profile function $Z$ (Fig.~\ref{fig:merge}c). After $Z$ is obtained, naively one would use $Z(\mathbf{r})$ as the height of the vehicle $z(\mathbf{r})$. However, for a vehicle with a finite size, the actual physical contacts between the wheels and the membrane occur near the circumference of the disk. Thus, the vehicle height $z(\mathbf{r})$ can be approximated by the average membrane height $Z(\mathbf{r})$ around the disk circumference ($z(\mathbf{r})\approx\langle Z(\mathbf{r}')\rangle_{|\mathbf{r}'-\mathbf{r}|=R_v}$, see Fig.~\ref{fig:membrane}c and S9 in SI). On our circular membrane, the analytical solution to $Z$ evaluated in $z$ yields the vertical position $z_i$ for vehicle $i$ with mass $m_i$ as:
\begin{eqnarray}
2\pi\lambda z_i &=& \frac{\pi}{2}(|\mathbf{r}_i|^2-R^2)+\frac{m_i}{\sigma}\log{\left(\frac{R_v R}{R^2-|\mathbf{r}_i|^2}\right)}\nonumber\\
&+&\frac{1}{\sigma}\sum_{j\neq i} m_j \left(\log{\frac{|\mathbf{r}_i-\mathbf{r}_j|}{|\mathbf{r}_i-\mathbf{r}'_j|}}-\log{\frac{|\mathbf{r}_j|}{R}}\right),
\end{eqnarray}

where $\mathbf{r}_i,\mathbf{r}'_i=(R/|\mathbf{r}_i|)^2 \mathbf{r}_i$ are the planar position of the $i$th vehicle and its image charge, $R$ and $R_v$ are the radii of the membrane and the vehicle, $\sigma$ is the area density of the membrane and $\lambda=v_m^2/P_0$ is a membrane constant. The three terms in the solution show respectively the contributions of the vehicle height field from the membrane, the weight of the vehicle of interest, and the other vehicles respectively.  The last term conceptually acts as an attractive potential (like the Newtonian gravitational potential), whose gradient generates a pairwise attractive force between the vehicles. 

With $z$ and therefore the acceleration as a function of the terrain gradient $-\nabla z$ at hand, one obtains the new position of the vehicles by integrating Eq.~(\ref{eq:eomGen1},\ref{eq:eomGen2}). This type of temporal updating is used to obtain the dynamics of binary systems under the Post-Newtonian approximation~\cite{Futamase2007}.

\section*{Attraction and cohesion dynamics in a two vehicle system}

\begin{SCfigure}[][ht!]
    \centering
    \includegraphics[width=0.45\textwidth]{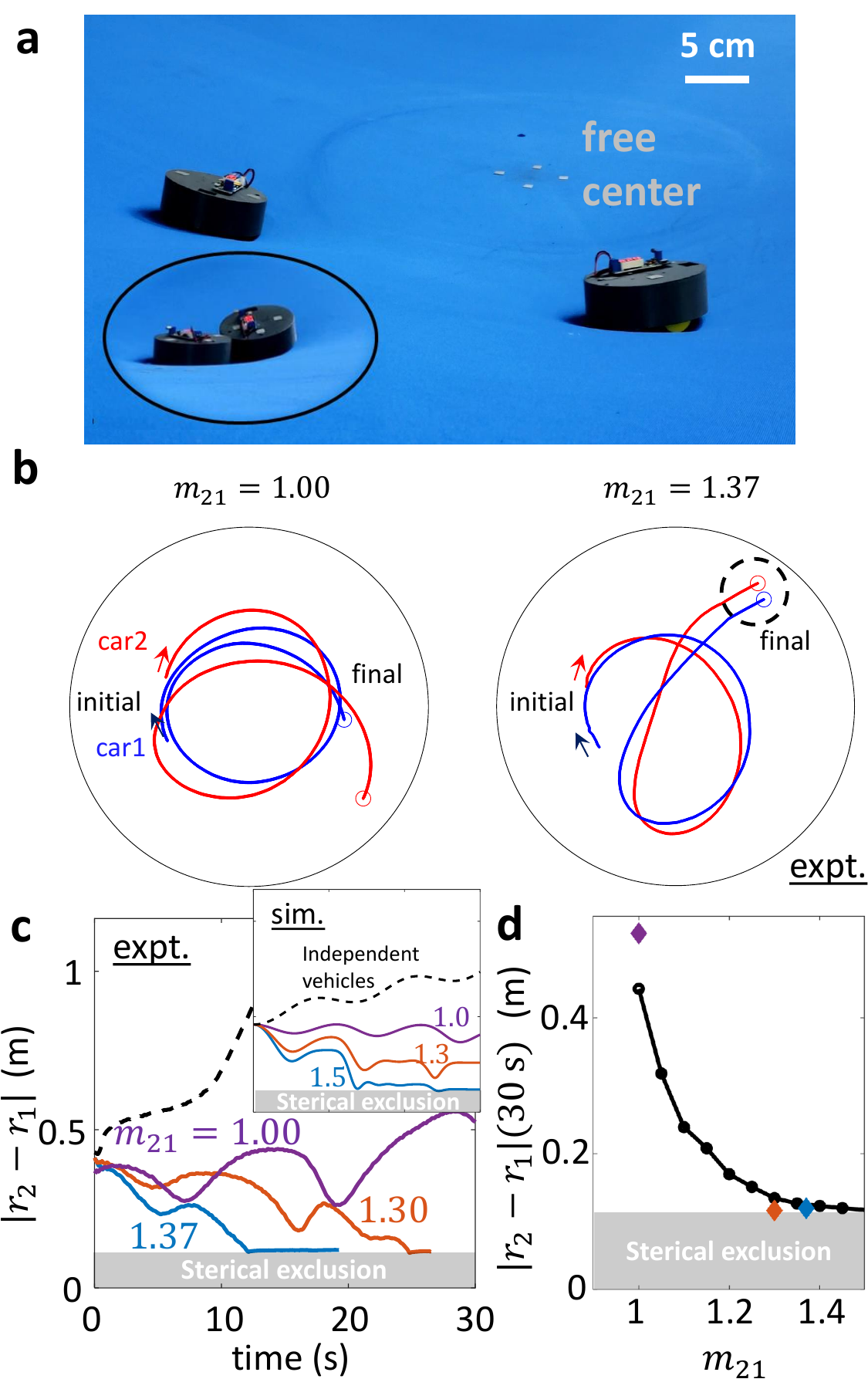}
    \caption{\textbf{Substrate deformation induced cohesion.} (a) Two vehicles moving on the elastic membrane and merge due to the substrate-mediated attraction. The initial azimuth angle between the vehicles is $45^\circ$. (b) Example trajectories of the two interacting vehicles  with different mass ratio ($\frac{m_2}{m_1}=m_{21}=1.00$ and $1.37$) for a duration of 30 sec. See Movie S5.mp4 for videos. (c) The evolution of the relative distance between the two interacting vehicles (solid) compared to the non-interacting case (dashed) where two vehicles with $m_{21}=1.30$ were released individually from the same initial condition. The time to merge is shortened by the increased masses of the leading vehicle (vehicle 2, $m_2$). On the contrary, the distance between two independent vehicles from the same initial condition is non-decaying. (d) Simulation shows the distance at $30$ s between the two vehicles decays with the mass ratio $m_{21}$. Experiment results from (c) are shown in diamonds.}
    \label{fig:merge}
 \end{SCfigure}

While a full systematic study of interaction dynamics for arbitrary initial conditions of the two vehicles is beyond the scope of the work, integration of the above multi-body dynamical model reveals that surprisingly the simulation does not predict strong attraction between two vehicles with the same mass at the same speed (unless they are started facing each other); experimental measurements of robot interaction are in accord with this prediction (see Fig. \ref{fig:merge}b,c) such that two equally massed vehicles will undergo many transits around the membrane without cohering. This is analogous to the eccentric Kozai-Lidov mechanism, where the eccentricity excitation reduces when the masses become equal to each other and decreases the merger rate \cite{Naoz16}.  In contrast, the simulation predicts that in a situation where one vehicle trails another (Fig. \ref{fig:merge}), increasing the mass of the leading vehicle can increase the attraction (Fig. \ref{fig:coupling}) and lead to vehicle merger (cohesion). 

To test the hypothesized merger enhancement with the increase of the leading vehicle's mass, we experimentally increased the mass of the leading mass vehicle (small weights were attached to the top of the vehicle without changing the center of mass), $m_2$, relative to the trailing vehicle, $m_1$ (characterized by the mass ratio $m_{21}=m_2/m_1$). For each experiment, both vehicles were placed at a radial distance of $60$~cm from the center with azimuthal separation $\psi = 45^\circ$ and both with a heading of $\theta=90^\circ$. Before each experiment, we set the speed of the two vehicles to $0.2$~m/s by manually adjusting voltage of the motors. Due to the finite voltage, the speed slightly drops ($< 10~\%$) when the separation between the two vehicles decreases. Fig.~\ref{fig:merge}b shows how the dynamics depend on the mass ratio.  When $m_{21}=1$, both vehicles  execute nearly-circular orbits (left panel) and generally do not merge in a short time (see S8 of SI). As $m_{21}$ is increased to $1.37$, the trailing vehicle eventually becomes ‘captured’ by the leading vehicle leading to an effective cohesion such that the vehicles collide and then continue to move together for the duration of the experiment (right panel).

To quantify the attraction and cohesion dynamics, we measured the Euclidean distance between the vehicles projected onto the horizontal plane, $|\mathbf{r}_1-\mathbf{r}_2|$, as a function of time. We find that the time to capture is reduced as the mass of the leading vehicle increases (Fig.~\ref{fig:merge}c). For instance, when  $m_{21} = 1.30$, it took around $25$~s for the trailing vehicle to become captured (i.e., the vehicles collide when $|\mathbf{r}_1-\mathbf{r}_2| = 2R_v$). When $m_{21} = 1.37$, this capture occurred significantly faster, with the vehicles colliding in about $12$~s. The coupling effects are highlighted by contrasting to the dynamics from independently conducted single-vehicle experiments, one with the initial conditions of the ``leading'' vehicle and the other with the initial conditions of the ``trailing'' vehicle. The distance evaluated from these two independent trajectories shows a non-decaying trend that differs from the cases with both vehicles on the membrane (dashed lines in Fig.~\ref{fig:merge}c). Simulations using the same setup as the experiments show qualitative match with the experiments and the distance between the two vehicles decreases with the mass ratio $m_{21}$ (Fig.\ref{fig:merge}d).

\section*{Controlling cohesion between multiple vehicles}

Given that unequal mass cars typically collide and cohere after some time, we next sought how we could \emph{actively} mitigate such attraction. As demonstrated above, reducing vehicle mass can lessen the local deformation field to reduce cohesion, but active variation in this parameter is challenging. Intuitively, one could also control the vehicle to increase speed when it nears a high curvature region, thereby allowing the robot to accelerate out of such a region. We note that such a strategy is interesting as the robots could avoid (or potentially enhance) aggregation solely via local field measurements alone. Such dynamics could be useful for future swarms of sensory challenged robots \cite{li2021programming} moving on highly deformable environments. 

\begin{SCfigure}[][ht!]
    \centering
    \includegraphics[width=0.5\textwidth]{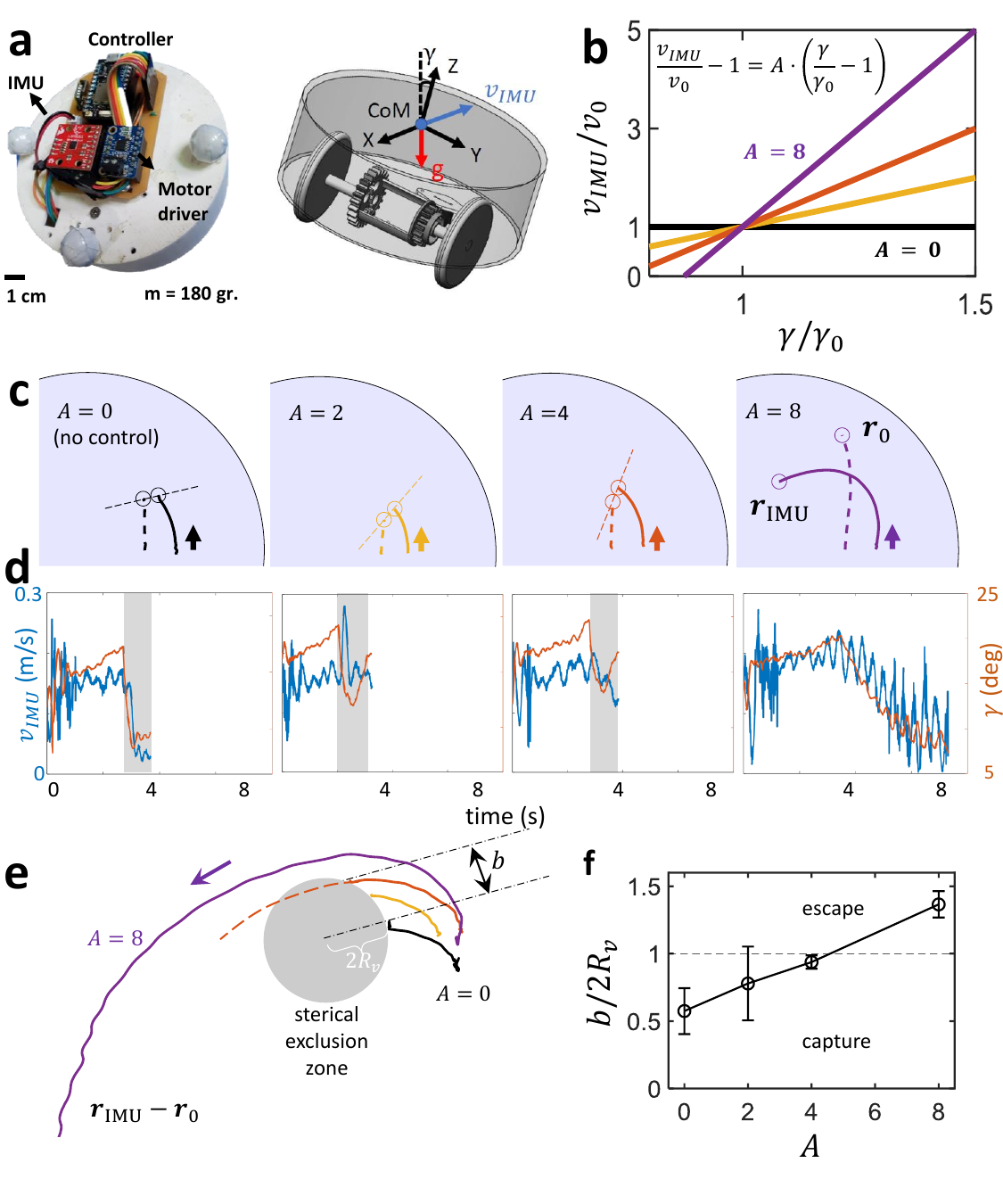}
    \caption{\textbf{Speed based on local tilt reduces substrate deformation induced cohesion.}
(a) A controller, IMU and DC motor driver are mounted on the speed-controlled vehicle ($m\approx$ 180 g) that changes speed according to the current tilt angle (right) with the same mechanics of the uncontrolled vehicle.
(b) Control scheme of the controlled vehicle. The speed $v$ increases with the tilt angle $\gamma$ to avoid collision. The control parameter $A$ increases from $0$ (black, no control) to $8$ (purple).
(c) shows the trajectories of the controlled vehicle (solid line) and the uncontrolled vehicle (dashed line) when different magnitudes of control are applied. The relative angle between two vehicles upon collision (dotted line) increases with $A$. See Movie S6.mp4 for movies.
(d) shows the evolution of the speed and tilt of the controlled vehicle corresponding to panel c. The shaded regions denote the collisions.
(e) shows the trajectories of the IMU-controlled vehicle in the frame of the uncontrolled vehicle. The sterical exclusion zone has a radius twice the radius of a vehicle. In an increasing order of control magnitude $A =0,2,4,8$, the trajectories gets further and further away from the uncontrolled vehicle with an increasing margin $b$. (f) Mean and standard deviation of $b/2R_v$ over 3 trials for different $A$ values. The vehicle eventually avoids the collision when $b/2R_v>1$.}
    \label{fig:escape}
 \end{SCfigure}

To test this cohesion mitigation strategy, we recall from the above section that as the distance between the two vehicles decreases, each `feels’ the membrane-induced deformation of the other more strongly and the tilt of both vehicles increases.  To measure local tilt angle $\gamma$ we added an IMU (Internal Measurement Unit) to the leading vehicle (Fig.~\ref{fig:escape}a) and implemented an adaptive speed controller that increased the speed of the leading vehicle as its measured $\gamma$, the angle of inclination from the gravity vector, increased in response to larger substrate deformations, from the IMU. Specifically, the speed of the leading vehicle was designated to be $(v_{\text{IMU}}-v_0)/v_0=A\cdot(\gamma-\gamma_0)/\gamma_0$, where $A$ sets the strength of the coupling between the leading vehicle and the local membrane deformation (Fig.~\ref{fig:escape}b).

The speed of the vehicle changes more quickly in response to the tilt when $A$ is larger. We varied $A$ from $0$ (no control; constant speed) to $8$ (speed sensitive to tilt angle) to probe the effects of the speed-tilt coupling strength on potential collisions with the trailing vehicle. Fig.~\ref{fig:escape}c shows the trajectories of the vehicles starting from the same initial conditions ($r_{\text{IMU}}(0)=0.6$ m, $r_{\text{passive}}(0)=0.4$ m, $\theta_{\text{IMU}}(0)=\theta_{\text{passive}}(0)=90^\circ$, $v_{\text{passive}}=0.11$ m/s, $v_{\text{IMU}}(0)=0.15$ m/s, and $\gamma_0=15^o$ ) for different $A$. From the recorded vehicle 3D position and orientation, we measured the speed and the tilt angle of the leading vehicle as a function of time; these measurements revealed that the controller generated the desired speed variation with tilt (Fig.~\ref{fig:escape}d). 

The robot's strategy, based solely on local curvature, leads to an ability to avoid collisions without knowing the location of the other vehicle. We observed that when $A$ was sufficiently large ($\geq 4$), the leading vehicle was able to successfully avoid collision. Fig.~\ref{fig:escape}e shows the relative trajectories of the controlled (leading) vehicle in the frame of the uncontrolled (trailing) vehicle ($\mathbf{r}_{\text{IMU}}-\mathbf{r}_{0}$). The steric exclusion zone (with radius equal to $2R_v$) around the uncontrolled vehicle identifies the collision area. If the controlled vehicle enters this area, then a collision with the uncontrolled vehicle has occurred. As $A$ increased, the margin, $b$, (i.e., the shortest distance between the controlled vehicle trajectory and the center of the uncontrolled vehicle) increased and eventually became larger than $2R_v$, indicating that the vehicles escaped the collision (Fig.~\ref{fig:escape}f). We note that the trajectory of the uncontrolled vehicle ended prematurely when a collision occurred; therefore, we fit it with an ellipse centered at the uncontrolled vehicle to extrapolate the margin $b$. 

The active cohesion mitigation strategy can be viewed in the spacetime framework by considering that the effect of the speed of the vehicle on the metric. This however requires derivation the metric functions analogous to Eq.\ref{eq:alpha1},\ref{eq:Phi1} for the more general case (Eq.\ref{eq:eomGen1},\ref{eq:eomGen2}) where vehicles can move on arbitrarily curved terrain. While we refer the reader to S7 in SI for details of the full derivation, we note here that for small vehicle accelerations, we obtain the temporal and spatial curvatures functions:
\begin{eqnarray}
\alpha^2 &=& E^2 (1-v^2 e^{-Cgz/v^2})\label{eq:alpha}\\
\Phi^2 &=& E^2 e^{-Cgz/v^2}(1-v^2 e^{-Cgz/v^2}).\label{eq:Phi}
\end{eqnarray}

The equations above reveal that the increase of speed $v$ makes the metric approach (conformal) flatness since the ratio between the temporal curvature $\alpha^2$ and spatial curvature $\Phi^2$ approaches unity with a higher speed.
The flatter spacetime implies smaller interaction between the vehicles. Thus our adopted strategy can be viewed as reduction of interaction by manipulating the effective spacetime metric.  A fruitful future direction could be to use such insights to develop controllers to mitigate effect from unexpected perturbation from the environment for the controlled agents such as robots.

\section*{Discussion \& Conclusions}

In this work, we studied the dynamics of active agents on an elastic substrate with interactions mediated solely by local and global deformation (curvature fields). Experimentally, we studied the dynamics of a single vehicle on a centrally curved elastic surface, revealing  nearly ubiquitous retrograde precessing orbits. We also studied the interaction of two vehicles on a relatively flat membrane with the feature that the time-dependent curvature fields of the vehicle reciprocally govern the robot's trajectories. We observed how increasing the mass ratio between the vehicles led to increased curvature field mediated cohesion. Finally, we developed control scheme for the multi-body system, which mitigated cohesion via speed increases of a robot upon measurement of its local tilt (an indirect measurements of the local curvature field). 

Theoretically, to understand the single vehicle orbital dynamics, we constructed a minimal mechanical model which agreed well with the experimental results and revealed that the vehicle did not simply follow geodesics of the membrane. Inspired by the resemblance of dynamics in our system to those in General Relativity (GR), we wondered if the GR framework could be of use to better understand features of the system. Remarkably, the active nature of the self-propelled robot allowed us to recast the robot's dynamics as geodesics of a test particle in an effective space-time. The spacetime framework revealed how aspects of the system (e.g. retrograde precessing orbits) were related to system parameters and allowed us to modify the vehicle mass to more closely mimic GR orbits (changing from retrograde to prograde). To understand the reciprocal curvature field mediated interaction two robot system, we first generalized the mapping to understand how individual vehicles responded to arbitrary curvature fields. We developed an equation for how the vehicles generated curvature by showing that the Poisson equation could approximate the shape of the membrane deformed by the masses on it. Solving the Poisson equation analytically demonstrated the origin of forces on the vehicles. Combining the equation of motion for a vehicle on a generalized deformed surface revealed the role of vehicle mass in attraction.

Our studies of active systems on elastic fields (which we believe are the first of their kind) and the spacetime framework could provide tools to robotic studies \cite{Hu2003,floyd2006novel,Suzuki2007,KohWood2015,Chen2018} of a broad class of physical\cite{bush2006walking,couder2005dynamical} systems that are capable of traversing complex, heterogeneous environments with static and dynamic structures by coupling their motion with the environment. For instance, the local curvature could be used as an information in addition to the more conventionally used environmental information such as vision  \cite{Chung2018,Weinstein2018,Sergiyenko2021} and stress sensing \cite{Mayya2019,Li2021,wang2021emergent} in swarm robotics. In this regard, we are following our previous work in which we demonstrated that use of ideas from modern physics can aid robotics-- for example the development of anticipatory control schemes to reduce robot scatter from heterogeneities; this is analogous to how mechanical diffraction patterns of legged and snake robots~\cite{schiebel2019mechanical,rieser2019dynamics} inspired design of collision interaction schemes \cite{qian2015anticipatory,GoldmanNote}. Swarms of robots could benefit from further discovery of principles by which agents can interact (e.g. like the so-called Cheerios effect \cite{vella2005cheerios}) and even communicate via environmental modifications\cite{liu2007towards}.  Practically, we believe that our system and framework could be used by robots with limited sensing and control, for example in lightweight water-walking robots \cite{zhao2012water,KohWood2015} or self-propelled microrobots~\cite{LiuStrano2020} swarming on fluid membranes~\cite{LiuStranoNote}. 

Finally, given the flexibility in construction and programming vehicles makes our system an attractive target to push toward a mechanical analog GR system; while superficially our system resembles the famous educational tool used to motivate Einstein's view of spacetime curvature influencing matter trajectories \cite{white2002shape, middleton2014circular,middleton2016elliptical}, unlike such systems which are \textit{not} good analogs of GR, the activeness allows the dynamics of the vehicle to be dictated by the curvature of space-time not just the curvature of space as in splittable space-times ($g_{tt}$ is constant)~\cite{price2016spatial}. Thus we posit that mechanical analog ``robophysical'' \cite{aguilar2016review,AydinBookChap} systems can complement existing fluid \cite{unruh1981experimental,patrick2018black}, condensed matter \cite{steinhauer2016observation}, atomic, and optical \cite{zhu2018elastic,philbin2008fiber,belgiorno2010hawking} analog gravity systems \cite{Barcelo2011} given the ability to create infinite types of spacetimes. As an example, it is possible to modify the setup of our vehicle orbiting a single depression and obtain paths that in the effective spacetime are exactly geodesics of a Schwarzschild black hole. To do this, one needs an additional degree of freedom, which allows the speed of the vehicle to depend on the radial distance. With that choice, it is possible to fix $k$ and $v$ in such a way that the metric functions in the effective spacetime are those of Schwarzschild in isotropic coordinates. Given the spiral-in orbits of the single vehicle with slight asymmetry due to the effective friction is reminiscent of the material's orbit in the frictional disk around the black holes of stellar mass, we posit the mechanical analog models could be useful to explore interesting GR phenomena like the other analogs. We might even generate analogies to wave-like systems \cite{couder2005dynamical,bush2015pilot,BushQuantum}, for example increasing the speed of the vehicle $v_m \sim v$ to be comparable to disturbance propagation (such that the membrane would follow the wave equation).

\section*{Materials and Methods}
\subsection*{Vehicle preparation}
The 3D printed self-propelling differential drive vehicle has a mass of $165$~g and diameter of $10$~cm. The vehicle has two active rear wheels ($d_w=4$~cm) that are connected via a LEGO Technic - differential gears 24-16 (which allows independent rotation of the wheels (separated by $3.6$~cm) and one front caster (Pololu ball caster with $3/8$'' metal ball) for stability. The active wheels are driven by a Pololu 120:1 mini plastic gearmotor (4.5 V,	120 RPM, 80 mA) that provides constant speed (adjusted by potentiometer), with a maximum of $0.32 $~m/s.  The robot is powered by lithium ion polymer battery (3.7~V, 500~mAh, from Adafruit).

The lighter vehicle is also 3D printed and has a mass of $50$~g and diameter of $7.5$~cm. To reduce the weight, we changed the design of the cap by cutting the unused sections that does not have any connectors, battery holder etc. We used Pololu wheels ($d = 3$~cm),  Pololu ball caster with 3/8'' plastic ball, Lithium Ion Polymer Battery - 3.7~V 150 mAh and 6~V, low-power, 0.36 A (Adafruit), metal micro metal gearmotor (Pololu) and 5V voltage regulator (Pololu). The differential mechanism is same as the heavy vehicle.

The controlled vehicle has the same mechanical structure as the uncontrolled vehicle. An IMU (SparkFun 9DoF IMU Breakout - LSM9DS1) is mounted on top of the robot. We control the speed of the DC motor by controlling the input voltage to the motor using Pulse Width Modulation (PWM) signal. The motor control module includes Particle Photon microcontroller and Adafruit DRV8833 DC motor driver breakout board. The speed of the motor is adjusted as a function of the tilt angle ($\gamma$, the angle of inclination from the gravity vector). The relation is given in Fig.~\ref{fig:escape}b. The tilt angle is calculated  as $\gamma = \arccos\frac{a_z}{\sqrt{a_x^2+a_y^2+a_z^2}}$ where $\mathbf{a}$ is the measured acceleration.

\subsection*{Membrane preparation}
Experimental setup consists of a trampoline (d = 2.5m, DICK'S Sporting Goods) covered with a 4-way stretchable spandex fabric (Rose Brand, 120" Spandex, NFR). 4-way stretchable refers to the fact that the strain-stress response in two perpendicular directions are the same, which provides maximum homogeneity. 

We adjusted the tension of the fabric homogeneously and then fixed the fabric to the metal frame using custom created holes and magnets. This adjustment allowed us to perform all the experiments under the same surface conditions. However, because we fixed the fabric manually, there is slight membrane anisotropy overall.  The custom made height controller attached to a steel disc (10 cm in diameter, Mcmaster) from the center of the setup with magnets. The controller includes Firgelli linear actuator (6inc stroke, 35 lbs) and Actuonix Linear Actuator Control Board that allow the control of the central height via LAC Software. We used Logitech webcam to take top view videos of the experiments and Optitrak motion capture system (with Flex 13 cameras and Motive software) to track the robot. 

\subsection*{Membrane characterization}
We used the motive capture system mentioned above to record the cross sections of the membrane for different central depressions and compared them with the theoretical solutions to a membrane the following Poisson equation with uniform load from the membrane self weight $\Delta Z = \lambda^{-1}$. The shape well follows the analytical solution $Z(r;R,R_0) = \frac{1}{4\lambda} r^2 + C_1(R,R_0) \log{r} + C_2(R,R_0)$ when proper membrane constant $\lambda= 6.5$ m is applied. This constant is also used for the multi-vehicle interaction computations. Check S8 in the SI for more details.

\subsection*{Trajectory collecting}
The position and orientation of the IR reflective markers on the robot are recorded with a motion capture system consisting of five Optitrack Flex 13 cameras with a resolution of $1.3$ MP/mm$^2$.

Considering the slight transient decay/expansion of orbit eccentricity (see S2 of the SI) and the membrane defects that shows up in the azimuthal every $2\pi$, we evaluate the precession by fitting the trajectory to a model $r(\varphi)=r_c + e^{-\varphi/\tau} \left(A_1 \cos{(\varphi+\varphi_1)}+ A_2 \cos{(\omega_{\text{prec}}\varphi+\varphi_2})\right)$ using the least square fitting with the experiment trajectories. Half lives of eccentricity are extracted from this model from the characteristic time $\tau$. 

\section*{Acknowledgment}
We thank Enes Aydin, Mariam A. Marzouk, Alia Gilbert, Maria Jose Serrato Gutierrez, Camila Dominguez, Steven Tarr for experimental assistance, Andrew Brown for computational assistance, Perrin Schiebel for the suggestion of using differential motor, Andras Karsai, Ram Avinery for proofreading, Zachary Goddard for creating early version of the vehicle, and Gary Gibbons for encouragement during early stages of the project. Funding for D.I.G, S.L, Y.O.A, and J.M.R provided by ARO and ARL MAST CTA; funding also provided by Dunn Family Professorship. P.L. supported by NSF Grants 1908042 1806580 1550461.

\bibliographystyle{unsrt}  

\bibliography{references}

\clearpage
\section*{Supplementary Materials}
\subsection*{S1. Vehicle Dynamics}
\begin{figure}[ht!]
  \centering
  \includegraphics[width=0.5\textwidth]{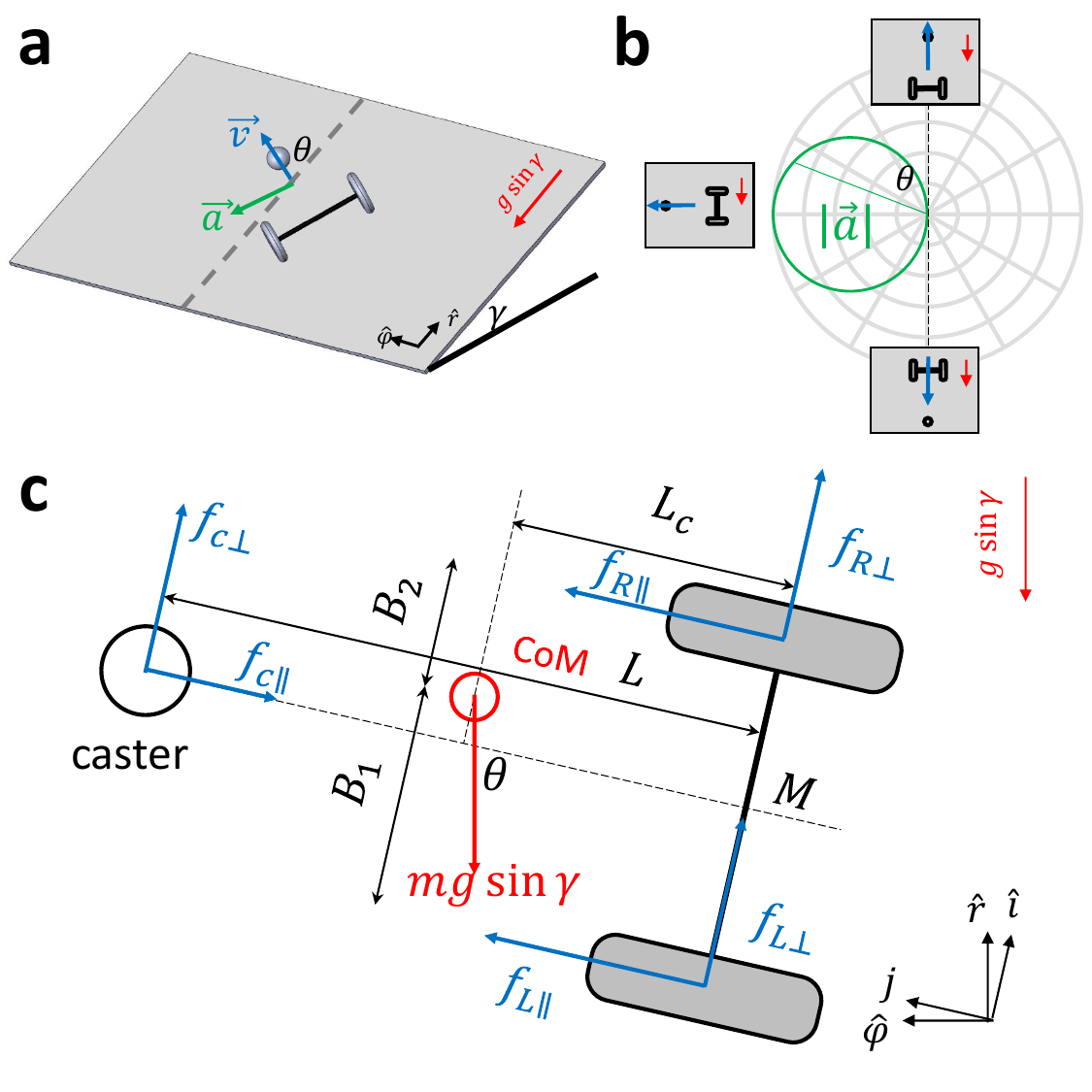}
  \caption{\textbf{Vehicle dynamics of the robotic vehicle.} (a) Modelling the dynamics of the vehicle on a slope with incline angle equalled to its current tilt $\gamma$. (b) The magnitude of the acceleration changes with the heading angle $\theta$ and vanishes when going along the gradient of the incline. (c) The force diagram of the vehicle.}
\label{fig:carMechanics}
\end{figure}

The dynamics of the vehicle on an incline with slope $\gamma$, which is a localized representation of substrate under the vehicle helps explain the acceleration's dependence on the heading $\theta$ and local tilting angle $\gamma$ (Fig.~\ref{fig:carMechanics}a) in experiments. On the incline, we denote the direction along the gravity as $\parallel$ and the direction perpendicular to it as $\perp$ so that the acceleration from the gravity field is $ a_{\perp}^g=0,\sim a_{\parallel}^g=g\sin{\gamma}$ (Fig.~\ref{fig:carMechanics}b). Considering this incline as a localized picture of the vehicle's immediate substrate, here $\hat{\perp}$ direction stands for the $\hat{\varphi}$ and $\hat{\parallel}$ direction stands for the $\hat{r}$.\\

Since the friction on the rolling caster is much smaller than the other friction forces, the vehicle rotates about the middle point of the wheel axis, $M$. The torque about $M$ consists of the frictions on the two wheels and the caster, as well as the gravity component in the plane. Since the two wheels are connected to a differential drive, the torques generated by the friction parallel to the wheel $f_{L\parallel},f_{R\parallel}$ are of same magnitude and opposite signs and therefore cancelled out. The torques generated by the friction perpendicular to the wheel are zero since the forces pass through $M$.

The non-zero torques left with us are the one generated by the gravity component in the plane and the friction from the caster $f_c$:
\begin{eqnarray}
    \tau &=& \left(\Delta B~\hat{i} + L_c~ \hat{j}\right) \times \nonumber\\
    && mg\sin{\gamma} (-\sin{\theta}~\hat{i} - \cos{\theta}~\hat{j})
    +L~\hat{j} \times f_{c\perp}~ \hat{i}\\
    &=& \left(mg \sin{\gamma} \left(-\Delta B \cos{\theta} + L_c \sin{\theta}\right) -f_{c\perp} L\right) \hat{k}
\end{eqnarray}
where $\Delta B \equiv \frac{1}{2}(B_2-B_1)$.

The moment of inertia of the vehicle with respect to $M$ is $I=I_{\text{vehicle}}+m(L^2 + \Delta B^2)$ where we approximate $I_{\text{vehicle}}=\frac{1}{2} m R_v^2$ with $R_v$ being the radius of the vehicle since the mass distribution is quite homogeneous. Therefore the magnitude of the acceleration of the center of mass is

\begin{eqnarray}
    a&=&\left|L_c~\hat{j} + \Delta B~\hat{i}\right| \cdot \beta\\
    &\approx& L_c \cdot\frac{\tau \cdot \hat{k}}{I}\\
    &=&\frac{mg\sin{\gamma}(L_c \sin{\theta}-\Delta B\cos{\theta})-f_c L}{\frac{1}{2}mR_v^2+mL_c^2+m\Delta B^2}L_c
\end{eqnarray}

For the ideal case that the center of mass is not biased to the left or right so that $B_1=B_2$, the acceleration is

\begin{eqnarray}
a&=&\frac{mgL_c \sin{\gamma}\sin{\theta}-f_{c}L}{\frac{1}{2}mR_v^2+mL_c^2}L_c\nonumber\\
&=&\frac{L_c^2}{\frac{1}{2} R_v^2+L_c^2} g\sin{\gamma}\sin{\theta}- \frac{f_{c}L}{\frac{1}{2}mR_v^2+mL_c^2}
\end{eqnarray}

When $\theta = \pi/2$ and $f_{c}$ being very small since this is a rolling friction, the acceleration projected onto the horizontal plane is
\begin{eqnarray}
a(\theta=\pi/2)\approx \frac{L_c^2}{\frac{1}{2}R_v^2+L_c^2} g\sin{\gamma}\cos{\gamma}
\end{eqnarray}

The actual numbers in the experiment $R_v=5~$cm, $L_c \approx 1~$cm give the theoretical prediction as
\begin{eqnarray}
a_{\text{theo}}(\theta=\pi/2) \approx 0.074~g\sin{\gamma} \cos{\gamma}
\end{eqnarray}
which is quite close to the experimental measurement
\begin{eqnarray}
a_{\text{expt}}(\theta=\pi/2) = (0.073 \pm 0.001)~g\sin{\gamma} \cos{\gamma}
\end{eqnarray}

In reality, there is always a small bias between $B_1$ and $B_2$, this small correction from the CoM (center of mass) offset that breaks the symmetry of acceleration with respect to the heading gives the attraction to the circular orbit and will is discussed in Section S3.

\begin{SCfigure}[][ht!]
    \centering
    \includegraphics[width=0.35\textwidth]{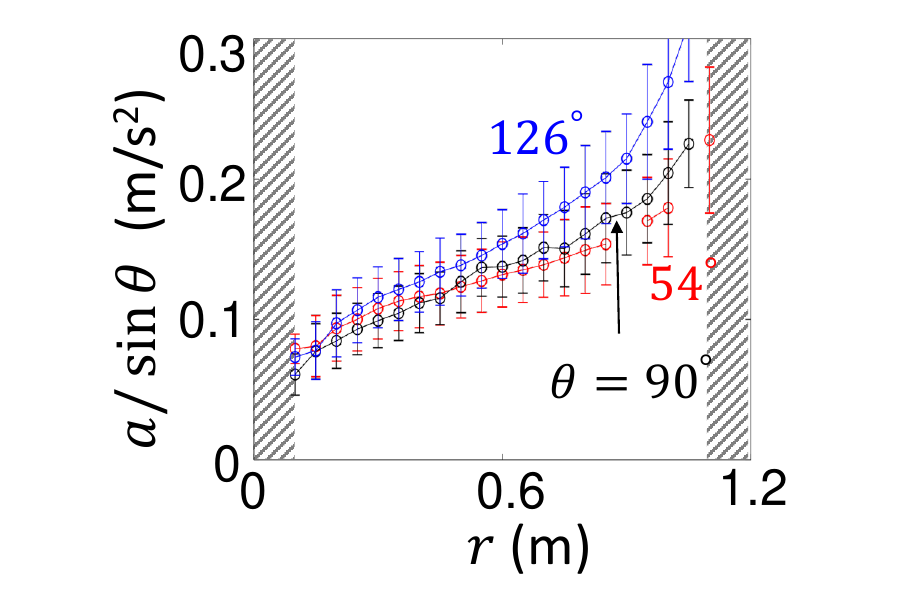}
    \caption{Plots of $k$ as a function of $r$ for various values of $\theta$ using $a/\sin{\theta}$. The gray shaded regions refer to regions which are forbidden due to steric exclusion.}
    \label{fig:headingDependence}
\end{SCfigure}

This bias is
\begin{eqnarray}
    a_{\text{bias}} = g\,\sin{\gamma}\cos{\theta}\, \frac{L_c \Delta B}{\frac{1}{2}R^2+L_c^2+\Delta B^2}
\end{eqnarray}
where $\Delta B$ can be measured by weighing the normal force on the left and right wheels and given by
\begin{eqnarray}
    \Delta B=\frac{L_w}{2}\frac{N_R-N_L}{N_R+N_L}
\end{eqnarray}
where $N_L, N_R$ are the normal forces on the two wheels and $L_w = 6~$cm. For an imbalance of $(N_R-N_L)/(N_R+N_L) \approx 20~\%$ thus $\Delta B \approx 0.6~$ cm. Thus, the maximum bias ($\theta = 0^\circ, 90^\circ$) when driving on a typical local slope of $\gamma = 10^{\circ}$ is $a_{\text{bias}} = 0.074~\text{m/s}^2$, which is about $40~\%$ of the maximum magnitude of the acceleration in the system. Fig.\ref{fig:headingDependence} shows how this bias causes the slight dependence on $\theta$.

\clearpage
\subsection*{S2. Transient Factor}
The transient behavior of some trajectories that decay into circular orbits is caused by the slight asymmetry in the mechanical structure that the center of mass (CoM) deviates slightly from the center-line. As shown in Section S1, such defected acceleration magnitude $|a|$ is given by
\begin{equation}
|a|=k(r)\cdot (\sin{\theta} +\epsilon \cdot \cos{\theta})
\end{equation}
where $\epsilon = \frac{L_c \Delta B}{\frac{1}{2}R^2+L_c^2+\Delta B^2} \ll 1$ increases with the CoM's deviation from the center-line being $\Delta B$.

Following the treatment shown in the main text, the polar equation of the trajectory is
\begin{eqnarray}
    r_{,\varphi\varphi}=\frac{2r_{,\varphi}^2}{r}+r-\tilde{k}(r)\cdot (r^2 +r_{,\varphi}^2)\nonumber\\
    -\epsilon \cdot \tilde{k}(r) \cdot (r_{,\varphi}r+\frac{r_{,\varphi}^3}{r})
\end{eqnarray}
where $\tilde{k}\equiv k/v^2$.

Let $r=r_c+\rho$ where $\rho$ is the perturbation and $r_c$ is the radius of the circular orbit that $k(r_c)=v^2/r_c$. After discarding the $O(\rho^2)$ terms, the differential equation is reduced to

\begin{eqnarray}
    \rho_{,\varphi\varphi}=-(1+r_c k'_c/k_c)\rho -\epsilon \rho_{,\varphi}
\end{eqnarray}
where $k_c\equiv k(r_c), k'_c\equiv k'(r_c)$.

The solution to this damped oscillator gives the solution as
\begin{eqnarray}
    \rho(\varphi)=\rho(0)~ \cos{\left(\sqrt{1+r_c k'_c/k_c - (\epsilon/2)^2}\varphi\right)}~ e^{-\epsilon\varphi/2} \label{eq:analApprx}
\end{eqnarray}
with an exponentially decaying envelope with a half-life $(2\log{2})/\epsilon$ that degrades with the bias; that is, the larger the imperfection is, the faster the trajectory is attracted a circular orbit.

On the other hand, when the vehicle has an acceleration bias towards the orbit direction, $\epsilon$ will be negative, then $\rho$ will expand and leads the orbit to either crash to the center or escape from the membrane. From this example with counterclockwise trajectory, we see that the orbit is attracted to a circular orbit when $\epsilon \propto (B_2-B_1)>0$, that is when the CoM is biased to the left wheel. The data listed in Section S1 shows an estimate $\epsilon \approx 0.043$, indicating a half life of $(2\log{2}/0.0043)\approx 22$ revolutions. This matches with our experimental observation.\\

In summary, a counterclockwise(clockwise) orbit will get attracted to a circular orbit when the CoM is biased to the left(right) while the eccentricity increases to escape or crash when the CoM is biased to the right(left).\\

\clearpage

\subsection*{S3. Trajectory resulted from active vehicle deviates from spatial-only geodesics}
To measure the spatial-only trajectory, we let the left and right wheel speed of the vehicle be the same. The spatial-only trajectory (blue) enabled by the same left and right wheel speeds is much straighter than that of an active vehicle responding the local gradient (red). 

\begin{figure}[ht!]
  \centering
  \includegraphics[width=0.5\textwidth]{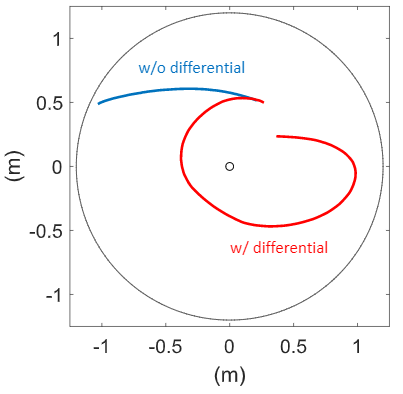}
  \caption{Comparison of the trajectories of the vehicle on the membrane when differential mechanism is applied and disabled. To disable the differential mechanism so that the two wheels are rigidly connected, the gears in the differential are glued.}
\label{fig:spatialGeodesics}
\end{figure}
\clearpage
\subsection*{S4. Metric Derivation (Axis-symmetric)}
In the axi-symmetric case, the geodesic equations for the metric $ds^2 = -\alpha^2 dt^2 +\Phi^2(\Psi^2 dr^2 + r^2 d\varphi^2)$ read
\begin{eqnarray}
&&\ringring{t} + \frac{(\alpha^2)'}{\alpha^2}\mathring{t}\mathring{r}= \frac{1}{\alpha^2}\left(\alpha^2\mathring{t}\right)^\circ = 0 \label{eq:geo1} \\
&&\ringring{\varphi} + \frac{(\Phi^2\,r^2)'}{\Phi^2\,r^2}\mathring{\varphi}\mathring{r}= \frac{1}{\Phi^2\,r^2}\left(\Phi^2\,r^2\mathring{\varphi}\right)^\circ =0 \label{eq:geo2}\\
&&\ringring{r} + \frac{(\alpha^2)'}{2\Phi^2\Psi^2}\mathring{t}^2+
\frac{(\Phi^2\Psi^2)'}{2\Phi^2\Psi^2}\mathring{r}^2 -\frac{(\Phi^2r^2)'}{2\Phi^2\Psi^2}\mathring{\varphi}^2 = 0\label{eq:geo3}
\end{eqnarray}
with $\lambda$ as an affine parameter and $\mathring{q}=dq/d\lambda,\ringring{q}=d^2q/d\lambda^2$. From Eqs.~(\ref{eq:geo1}) and (\ref{eq:geo2}), 
we have that
\begin{eqnarray}
\alpha^2\mathring{t} = E = \text{constant}, \label{eq:E}\\
\Phi^2 r^2\mathring{\varphi} = L = \text{constant},\label{eq:L}
\end{eqnarray}
both a consequence that conservation of energy and angular momentum holds.

With the help of $\mathring{q}=(dq/dt)(dt/d\lambda)=\mathring{t}\dot{q}$, the geodesic equations can be rewritten as
\begin{eqnarray}
\ddot{\varphi} + \frac{2\dot r \dot\varphi}{r} &=&  \left[\frac{(\alpha^2)'}{\alpha^2} - \frac{(\Phi^2)'}{\Phi^2}\right]\dot r\,\dot\varphi\label{eq:GGeo1} \\
\ddot{r} - \frac{r \dot{\varphi}^2}{\Psi^2} + \frac{\Psi'}{\Psi}\dot r^2&=&  \left[\frac{(\alpha^2)'}{\alpha^2} - \frac{(\Phi^2)'}{\Phi^2} \right]\dot r^2\nonumber\\
&+& \frac{1}{2\,\Phi^2\Psi^2}\left[ (\Phi^2)'v^2-(\alpha^2)'\right]\,. \label{eq:GGeo2}
\end{eqnarray}

We recall that the equations of motion of  the vehicle are
\begin{eqnarray}
 \ddot{\varphi} + \frac{2\,\dot r \,\dot\varphi}{r} &=&\frac{k\Psi}{v^2}\label{eq:ref1} \dot{r}\dot{\varphi}\\
\ddot{r} - \frac{r \,\dot{\varphi}^2}{\Psi^2} + \frac{\Psi'}{\Psi}\dot r^2 &=& \frac{k\Psi}{v^2}\dot{r}^2-\frac{k}{\Psi}\label{eq:ref2}
\end{eqnarray}

Equating the right hand sides of Eqs.~(\ref{eq:GGeo1}) and (\ref{eq:GGeo2}) with those of (\ref{eq:ref1}) and (\ref{eq:ref2}) respectively yields 
\begin{eqnarray}
\frac{(\alpha^2)'}{\alpha^2} &=& \frac{k\Psi}{v^2}\left[\frac{\Phi^2v^2}{\alpha^2-\Phi^2v^2}\right] \label{eq:k1}\\
\frac{(\Phi^2)'}{\Phi^2} &=& \frac{k\Psi}{v^2}\left[\frac{2\Phi^2v^2-\alpha^2}{\alpha^2-\Phi^2v^2}\right]\,.\label{eq:k2}
\end{eqnarray}
which after integration one gets
\begin{eqnarray}
\alpha^2&=&-\frac{1}{C_1 v^2} + C_2 \cdot e^{-K/v^2}\\
\Phi^2&=& \frac{\alpha^2}{v^2}+C_1 (\alpha^2)^2
\label{eq:curvature}
\end{eqnarray}
where $K=K(r) \equiv \int_0^r k(s) \Psi(s) ds$.

To determine the two constants $C_1$ and $C_2$, we need the help of the two following conditions.

\subsection*{S5. Normalization condition}
Noting the form of the normalization condition $-1 = -\alpha^2 \mathring{t}^2 + \Phi^2(\Psi^2\mathring{r}^2 + r^2 \mathring{\varphi}^2)$, we see writing the $\mathring{t}^2$ as a function of $r$ would be helpful.
\begin{eqnarray}
  \mathring{r}^2 &=& \left(\frac{E}{\alpha^2}\dot{r}\right)^2\nonumber\\
  &=& \frac{E^2}{(\alpha^2)^2}\frac{1}{\Psi^2}(v^2-r^2\dot{\varphi}^2)\nonumber\\
  &=&\frac{E^2}{(\alpha^2)^2}\frac{1}{\Psi^2}\left[v^2-r^2\left(\frac{\alpha^2}{E}\mathring{\varphi}\right)^2\right]\nonumber\\
  &=&\frac{E^2}{(\alpha^2)^2}\frac{1}{\Psi^2}\left[v^2-r^2\left(\frac{\alpha^2}{E}\frac{L}{\Phi^2 r^2}\right)^2\right]
\end{eqnarray}
Plug this into the normalization condition, we have
\begin{eqnarray}
-1 &=& -\alpha^2 \mathring{t}^2 + \Phi^2\left(\Psi^2\mathring{r}^2 + r^2 \mathring{\varphi}^2\right)\nonumber\\
-1 &=& -\alpha^2 (E/\alpha^2)^2 \nonumber\\
&&+ \Phi^2\left[\Psi^2(\frac{E\dot{r}}{\alpha^2})^2 + r^2 (\frac{L}{\Phi^2 r^2})^2\right]\nonumber\\
  -1&=&-\frac{E^2}{\alpha^2} + \frac{\Phi^2 E^2 v^2}{(\alpha^2)^2}
\end{eqnarray}
Plug in the $\alpha^2$ and $\Phi^2$ derived earlier (Eq.\ref{eq:curvature}), we have
\begin{eqnarray}
  -\frac{1}{E^2}=C_1 v^2
\end{eqnarray}
Therefore $C_1=-\frac{1}{v^2 E^2}$.\\

\subsection*{S6. Flat Spacetime Limit}
When the force vanishes as $k(r)=0$, we should have $\alpha^2=\Phi^2$ so that the metric is flat and the vehicle will go straight.

$k(r)=0$ indicates $K(r)=\int_{s=0}^r k(s)\Psi(s) = 0$. Setting the under-limit of the integral of $k$ zero without the loss of generality since otherwise it will be absorbed by $C_2$. This limit reduces the metric to

\begin{align*} 
\alpha_0^2 &=-\frac{1}{C_1 v^2} + C_2 \label{eq:con21} \\
\Phi_0^2 &= \frac{\alpha_0^2}{v^2}+C_1 (\alpha_0^2)^2 
\end{align*}

Equate the above two equations using $\alpha_0^2=\Phi_0^2$, we have
\begin{align*}
  \alpha_0^2 &= \frac{\alpha_0^2}{v^2}+C_1 (\alpha_0^2)^2\\
  1 &= \frac{1}{v^2} +C_1(-\frac{1}{C_1 v^2}+C_2)\\
  C_1C_2&=1
\end{align*}

With the above pieced together, we have
\begin{eqnarray}
\alpha^2 &=& E^2(1-v^2 e^{-K/v^2})\\
\Phi^2&=&E^2 e^{-K/v^2} (1-v^2 e^{-K/v^2})
\end{eqnarray}

Plug the derived metric into Eq.\ref{eq:E},\ref{eq:L}, the effective angular momentum $\ell$ is therefore
\begin{eqnarray}
\ell \equiv \frac{L}{E} &=& e^{-K(r_0)/v^2}r_0 \cdot v
\end{eqnarray}
The maximum of $\ell$ is obtained at $r_0$ that
\begin{eqnarray}
\frac{\partial \ell}{\partial r_0} &=&e^{-K(r_0)/v^2}\left(1-\frac{r_0k(r_0)}{v^2}\right),
\end{eqnarray}
showing $r_0$ coincides with the circular orbit radius $r_c$ such that $k(r_c)=v^2/r_c$.
\clearpage
\subsection*{S7. Metric Derivation (General)}
Assuming $|\nabla z|\ll 1$ such that $\Psi^2 \approx 1$, for the general metric $ds^2=-\alpha(x,y,t)^2 dt^2 + \Phi(x,y,t)^2 (dx^2+dy^2)$, the connections $\Gamma_{bc}^a=\frac{1}{2}g^{ad}(g_{bd,c}+g_{cd,b}-g_{bc,d})$ are
\begin{eqnarray}
\begin{aligned}[c]
\Gamma_{tt}^t&=\frac{1}{2}\frac{(\alpha^2)^.}{\alpha^2}=(\log{\alpha})^\cdot\\
\Gamma_{ti}^t&=\frac{1}{2}\frac{(\alpha^2)_{,i}}{\alpha^2}\\
\Gamma_{ij}^t&=\frac{1}{2}\frac{(\Phi^2)^.}{\alpha^2}\eta_{ij}
\end{aligned}
\quad
\begin{aligned}[c]
\Gamma_{tt}^i&=\frac{1}{2}\frac{(\alpha^2)_{,i}}{\Phi^2}\\
\Gamma_{jt}^i&=\frac{1}{2}\frac{(\Phi^2)_{,i}}{\Phi^2}\delta_j^i\\
\Gamma_{jk}^i&=\frac{1}{2}\frac{(\Phi^2)_{,k}\delta_j^i+(\Phi^2)_{,j} \delta_k^i - (\Phi^2)_{,i}\eta_{jk}}{\Phi^2}
\end{aligned}
\end{eqnarray}
where $q^\cdot \equiv dq/dt$.

The geodesic equations $\ringring{x}^a+\Gamma_{bc}^a \mathring{x}^b\mathring{x}^c=0$ are
\begin{eqnarray}
\ringring{t}+(\log{\alpha})^\cdot \mathring{t}^2 +2(\log{\alpha})_{,i} \mathring{t}\mathring{x}^i+\frac{1}{2}\frac{(\Phi^2)^\cdot}{\alpha^2}\eta_{ij}\mathring{x}^i\mathring{x}^j=0\label{eq:geodesicsGen1}\\
\ringring{x}^i+\frac{1}{2}\frac{(\alpha^2)_{,i}}{\Phi^2}\mathring{t}^2 +2(\log{\Phi})^\cdot \mathring{t}\mathring{x}^i+2(\log{\Phi})_{,k} \mathring{x}^k \mathring{x}^i\nonumber\\
- (\log{\Phi})_{,i} \mathring{x}^k \mathring{x}_k=0\label{eq:geodesicsGen2}
\end{eqnarray}

Then we change the variable from proper time to time. Using the facts that $\mathring{q}=(dq/dt)(dt/d\lambda)=\mathring{t}\dot{q}$ and $\dot{x}^i\dot{x}_i=v^2$, Eq.\ref{eq:geodesicsGen2} can be rewritten as
\begin{eqnarray}
\frac{\ringring{x}^i}{\mathring{t}^2}+\frac{(\alpha^2)_{,i}}{2\Phi^2}+2(\log{\Phi})^\cdot \dot{x}^i +2(\log{\Phi})_{,k}\dot{x}^k\dot{x}^i\nonumber\\
-(\log{\Phi})_{,i}v^2=0
\end{eqnarray}

Note that
\begin{eqnarray}
\frac{\ringring{x}^i}{\mathring{t}^2}=\frac{(\mathring{t}\dot{x}^i)^\circ}{\mathring{t}}=\ddot{x}^i+\frac{\ringring{t}}{\mathring{t}}\dot{x}^i
\end{eqnarray}

and Eq.\ref{eq:geodesicsGen1} can be rewritten as
\begin{eqnarray}
\frac{\ringring{t}}{\mathring{t}^2}+(\log{\alpha})^\cdot+2(\log{\alpha})_{,i}\dot{x}^i+\frac{1}{2}\alpha^{-2}(\Phi^2)^\cdot v^2=0,
\end{eqnarray}

The geodesic equations in the lab frame are
\begin{eqnarray}
\ddot{x}^i=&-\left[\log{(\frac{\alpha^2}{\Phi^2})}\right]_{,j}\dot{x}^j\dot{x}^i+(\log{\Phi})_{,i}v^2-\frac{\alpha^2}{\Phi^2}(\log{\alpha})_{,i}\nonumber\\
&+x^i\left[(\log{\alpha})^\cdot+v^2(\log{\Phi})^\cdot \left(\frac{\Phi^2}{\alpha^2}-2\right)\right]\label{eq:geodesicsGen}
\end{eqnarray}

To match the geodesic equations Eqn.\ref{eq:geodesicsGen} to the general equations of motion for a constant-speed agent
\begin{eqnarray}
\ddot{x}&=&C\,g\, \dot{y}\,(d_x \dot{y}-d_y \dot{x})/v^2\\
\ddot{y}&=&-C\,g\, \dot{x}\,(d_x \dot{y}-d_y \dot{x})/v^2\,,
\end{eqnarray}
where $d_i = - \nabla_i z$ with $i = x,y$.

It requires
\begin{eqnarray}
-\frac{Cg}{v^2}d_i&=&\frac{(\alpha)^2_{,i}}{\alpha^2}-\frac{(\Phi^2)_{,i}}{\Phi^2}\\
0&=&-\frac{(\alpha^2)_{,i}}{2\Phi^2}+\left(\frac{(\alpha^2)_{,i}}{\alpha^2}-\frac{(\Phi^2)_{,i}}{2\Phi^2}\right)v^2\\
0&=&(\log{\alpha})^\cdot+v^2(\log{\Phi})^\cdot \left(\frac{\Phi^2}{\alpha^2}-2\right)\label{eq:metricConstraint}
\end{eqnarray}

If we consider the stationary metric that $\alpha$ and $\Phi$ are time-independent, then Eq.\ref{eq:metricConstraint} is met.

It can be checked that the solution is
\begin{eqnarray}
\log{(\alpha^2)}&=&C_0+\log{(1-v^2 e^{-Cgz/v^2})}\\
\log{(\Phi^2)}&=&C_0+z+\log{(1-v^2 e^{-Cgz/v^2})}.
\end{eqnarray}

Consequently,
\begin{eqnarray}
\alpha^2 &=& E^2 (1-v^2 e^{-Cgz/v^2})\\
\Phi^2 &=& E^2 e^{-Cgz/v^2} (1-v^2 e^{-Cgz/v^2})
\end{eqnarray}
\clearpage
\subsection*{S8. Membrane Measurement}
\subsection*{S8.1 Membrane constant}
To model the membrane deformation, we consider a free circular membrane with radius $R$ only deformed by its self weight and pressed by a cap in the center with depth $D$ and cap radius $R_0<R$. When the load from self weight is uniform, the height of the membrane $z$ follows
\begin{equation} \label{eq:PoissonEqnOrig}
\Delta Z = \lambda^{-1}
\end {equation}
where $\lambda$ absorbed the elasticity and the mass density.

Applying the axi-symmetry ($\partial Z / \partial \varphi = 0$) and boundary conditions $Z(R)=0, Z(R_0)=-D$ for a membrane without a load such as the robotic vehicle, the general solution to a membrane deformed by only self weight is

\begin{equation}
Z(r) = \frac{1}{4\lambda} r^2 + C_1 \log{r} + C_2 \label{eq:freeSln}
\end {equation}

where
\begin{eqnarray}
C_1 &=& \frac{D-\frac{1}{4\lambda}(R^2-R_0^2)}{\log{(R/R_0)}},\\
C_2 &=& \frac{\frac{1}{4\lambda}(R^2 \log{R_0}-R_0^2 \log{R})-D \log{R}}{\log{(R/R_0)}}
\end{eqnarray}

\begin{figure}[ht!]
  \centering
  \includegraphics[width=0.9\textwidth]{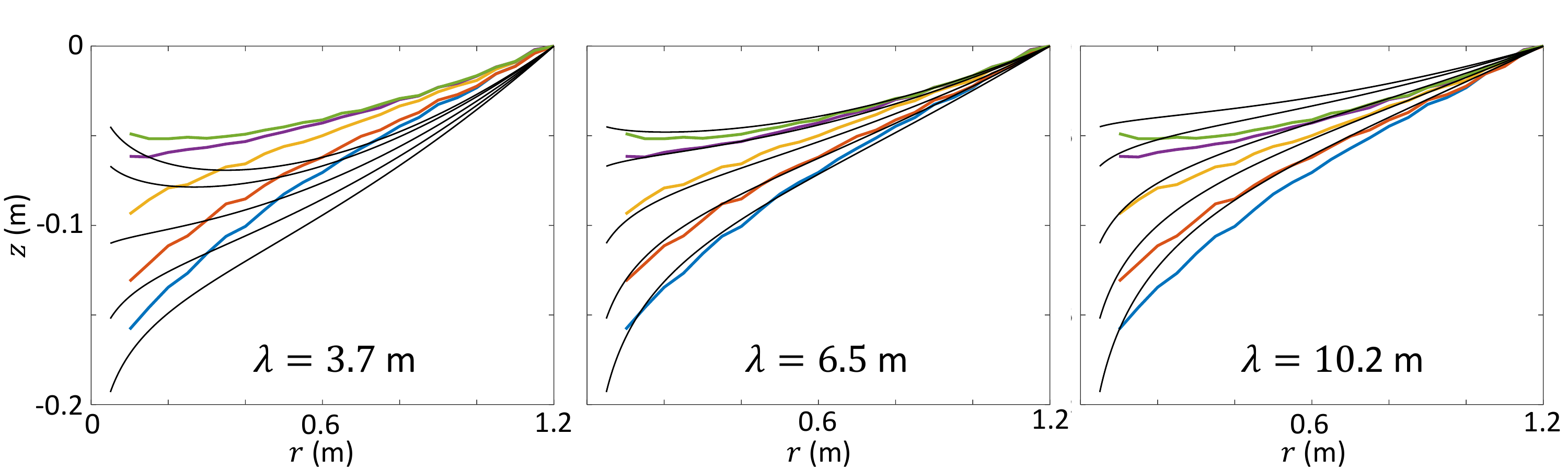}
  \caption{\textbf{Membrane constant measurement:} The black lines show the radial profiles of the free membrane from Poisson equation Eq.\ref{eq:freeSln}. The colored lines show the measurement from experiments.}
  \label{fig:freeMembrane}
\end{figure}

We measured the cross sections of the membrane with various central depressions $D$'s and compare them with solution Eq.\ref{eq:freeSln} for various $\lambda$. The value of $\lambda$ is chosen such that the solutions match with experiments the best. In our setup, $\lambda$ is measured to be $ 6.5$ m (Fig.\ref{fig:freeMembrane}).
\clearpage
\subsection*{S8.2 Membrane isotropy}
Ideally, the height of the membrane at a particular radius should be the same for any azimuthal angle in terms of the axi-symmetry. To understand how the membrane deviates from the ideal, the variation of this height is evaluated with the data taken from the optic tracking cameras for three different central depressions. The variation is found to be smaller than $5\%$ of the central depression.
\begin{figure}[ht!]
  \centering
  \includegraphics[width=0.9\textwidth]{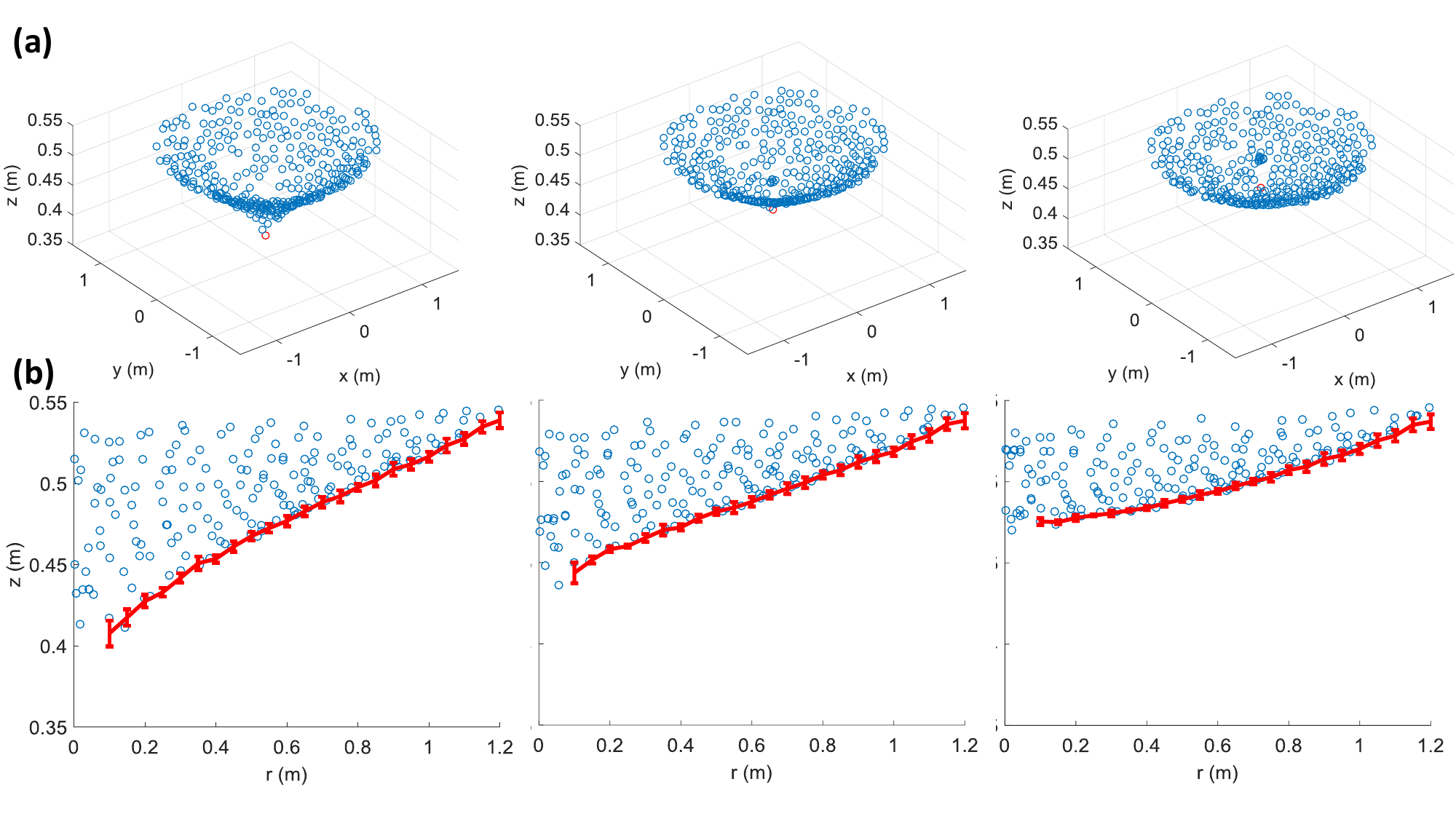}
  \caption{\textbf{Shapes of the membrane with different central depressions.} (a) The perspective views of the membrane profile measured from the optical tracking system. (b) The red curves show the heights averaged over the azimuthal angles.}
  \label{fig:membraneShape}
\end{figure}

\clearpage
\subsection*{S9. Analytic solution to the membrane}
As shown in the previous section, the deformation of the membrane by its self weight can be well characterized by $\Delta Z = \lambda^{-1}$. To model the additional load from the vehicles besides the weight of the membrane itself, we evaluate the area density of vehicle and scaled it by that of the membrane so that $\Delta Z = \lambda^{-1} (1+\tilde{P})$ with $\tilde{P}=\sigma_v/\sigma$ where $\sigma_v$ and $\sigma$ are the density of the vehicle and the membrane ($137$ g/m$^2$) respectively. For simplicity, we assume the load is a uniform distribution on a disc centered at the $i$th vehicle's position $\mathbf{r}_i$ and with the radius of the vehicle $R_v$ so that $\sigma_{v,i}=\frac{m_i}{\pi R_v^2} \mathbb{1}(\mathbf{r}\in \Omega_i)$ and $\sigma_v = \sum_i\sigma_{v,i}$ where $\Omega_i = \{\mathbf{r}:|\mathbf{r}-\mathbf{r}_i|<R_v\}$.

To solve the Poisson equation, we integrate the Green function $G(\mathbf{r},\mathbf{s})$ of Poisson equation with the source.

\begin{eqnarray}
\lambda Z(\mathbf{r})&=&\int G(\mathbf{r},\mathbf{s}) (1+\tilde{P}(\mathbf{s})) d\mathbf{s}^2\\
&=&\int G(\mathbf{r},\mathbf{s})d\mathbf{s}^2 + \frac{1}{\sigma}\sum_i \int_{\Omega_i} G(\mathbf{r},\mathbf{s}) \sigma_{v,i}(\mathbf{s}) d\mathbf{s}^2\\
&\equiv& I_1 + I_2
\end{eqnarray}

where the Green function on a disc with radius $R$ is
\begin{eqnarray}
G(\mathbf{r},\mathbf{s})&=&\frac{1}{2\pi}\log{|\mathbf{r}-\mathbf{s}|}\nonumber\\
&-&\frac{1}{2\pi}\log{\left(\frac{|\mathbf{s}|}{R}\cdot\left|\mathbf{r}-R^2 \frac{\mathbf{s}}{|\mathbf{s}|^2}\right|\right)}\\
G(\mathbf{r},\mathbf{0})&=&\frac{1}{2\pi}\log{|\mathbf{r}|}-\frac{1}{2\pi}\log{R}
\end{eqnarray}

Let us consider a field point that is not covered by the vehicles $\mathbf{r}\notin \cup_i \Omega_i$. $I_1$ is the solution to the case with uniform load that $I_1=\frac{1}{4}(|\mathbf{r}|^2-R^2)$. For $I_2$, the source is effectively a point source since the field point is outside the source, so
\begin{eqnarray}
I_2 &=& \frac{1}{\sigma}\sum_i \int_{\Omega_i} G(\mathbf{r},\mathbf{s}) \frac{m_i}{\pi R_v^2}\pi R_v^2\delta(\mathbf{s}-\mathbf{r}_i) d\mathbf{s}^2\\ &=&\frac{1}{\sigma}\sum_i m_i G(\mathbf{r},\mathbf{r}_i)
\end{eqnarray}

Up till so far, we have solved the shape of the membrane $Z(\mathbf{r})$. Next, we evaluate the height of the $i$th vehicle. Since the vehicle is not a point object, we average the membrane height $Z$ on the rim of the vehicle to approximate the height of the vehicle $z_i$.
\begin{eqnarray}
z_i&=&\langle Z\rangle_{\partial\Omega_i}\\
\lambda z_i&=&\langle I_1+I_2\rangle = \langle I_1\rangle + \langle I_2\rangle
\end{eqnarray}

$\langle I_1 \rangle$ is contributed by the self weight of the entire membrane so that we approximate it by just the value at the center of the vehicle $\mathbf{r}_i$: $\langle I_1 \rangle = \frac{1}{4}(|\mathbf{r}_i|^2-R^2)$.

For $\langle I_2 \rangle$, there are two different types of contributions. The first ones are the patches of domain from the vehicles other than the $i$th vehicle, the one of concern that contribute as far field. The second type is the contribution from the load of vehicle $i$ itself.

For the first type, we still use the point source approximation:
\begin{eqnarray}
\langle I_{2,j\neq i} \rangle = \frac{m_j}{\sigma}G(\mathbf{r}_i,\mathbf{r}_j)
\end{eqnarray}

For the second type:
\begin{eqnarray}
\langle I_{2,i} \rangle &=& \frac{m_i}{\sigma} \langle G(\mathbf{r},\mathbf{r}_i)\rangle_{\mathbf{r}\in\Omega_i}\\
&=&\frac{m_i}{2\pi\sigma}\left(\langle \log{|\mathbf{r}-\mathbf{r}_i|}\rangle-\left\langle\log{\left(\frac{|\mathbf{r}_i|}{R}\cdot\left|\mathbf{r}-R^2 \frac{\mathbf{r}_i}{|\mathbf{r}_i|^2}\right|\right)}\right\rangle\right)\nonumber\\
&=&\frac{m_i}{2\pi\sigma}\left(\log{R_v}-\log{\left(\frac{|\mathbf{r}_i|}{R}\cdot\left|\mathbf{r}_i-R^2 \frac{\mathbf{r}_i}{|\mathbf{r}_i|^2}\right|\right)}\right)\nonumber\\
&=&\frac{m_i}{2\pi\sigma}\log{\left(\frac{R_vR}{R^2-|\mathbf{r}_i|^2}\right)}
\end{eqnarray}

Piecing all these terms together, we arrive at the $z$ position of the $i$th vehicle is
\begin{eqnarray}
2\pi\lambda z_i&=& \frac{\pi}{2}(|\mathbf{r}_i|^2-R^2)+\frac{m_i}{\sigma}\log{\left(\frac{R_v R}{R^2-|\mathbf{r}_i|^2}\right)}\nonumber\\
&+&\frac{1}{\sigma}\sum_{j\neq i} m_j \left(\log{\frac{|\mathbf{r}_i-\mathbf{r}_j|}{|\mathbf{r}_i-\mathbf{r}'_j|}}-\log{\frac{|\mathbf{r}_j|}{R}}\right)\label{eq:analSlnMem}
\end{eqnarray}

where $\mathbf{r}'=(R/|\mathbf{r}|)^2 \mathbf{r}$ is conventionally regarded as the position of the image charge. $\mathbf{r}_j$'s are the positions of the other vehicles.

\begin{figure}[ht!]
  \centering
  \includegraphics[width=0.8\textwidth]{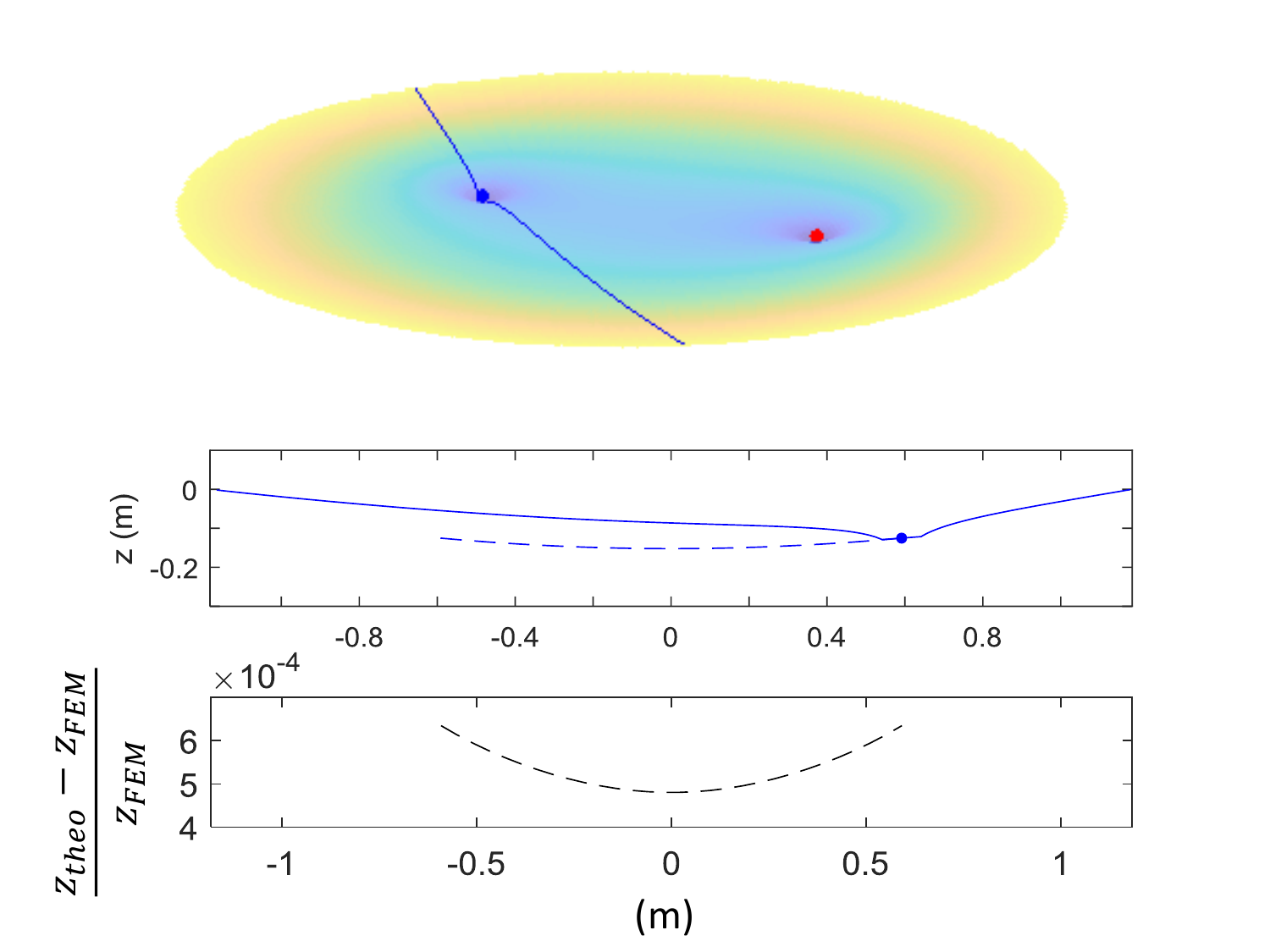}
  \caption{\textbf{Numerical verification of the analytical solution:} We show a test with the blue vehicle put at different $y$ positions while the $x$ position is fixed ($0.2$ m). The solid blue line shows the membrane shape and the dotted line shows the vertical position of the vehicle $z$ when placed at different positions. The bottom panel shows the relative error of $z$ between the analytical (Eq.\ref{eq:analSlnMem}) and numerical (FEM) solution.}
  \label{fig:theoVerification}
\end{figure}

Despite the fact that some approximations are made, the analytical solution matches with the numerical result (FEM) with a relative error smaller than $10^{-3}$ (Fig.\ref{fig:theoVerification}).

\newpage
\subsection*{S10. Dynamics of two vehicles with the same mass}

\begin{figure}[ht!]
  \centering
  \includegraphics[width=0.8\textwidth]{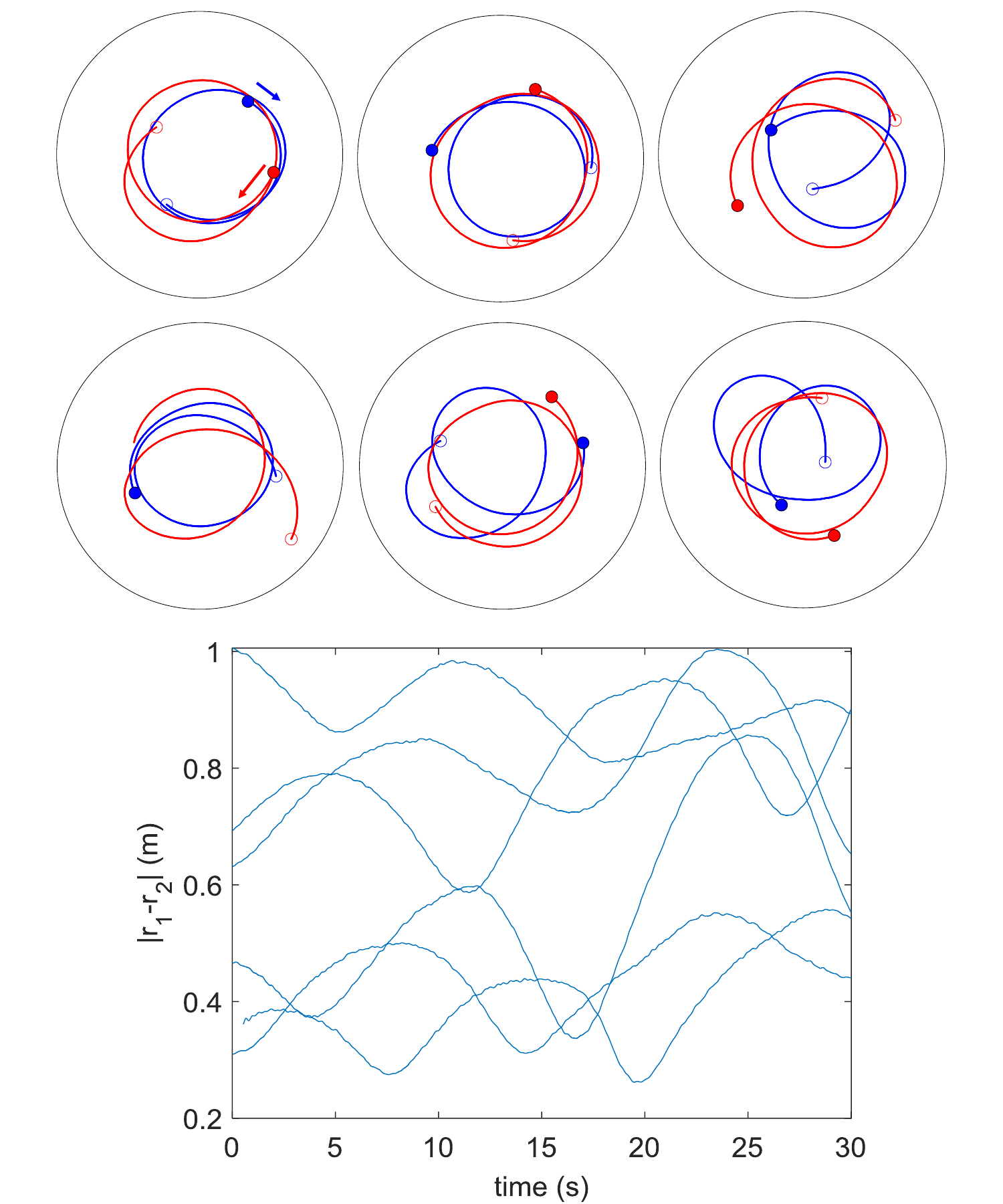}
  \caption{\textbf{Dynamics of two vehicles with the same mass} (a) Trajectories of vehicles with the same mass started at different initial conditions. (b) The relative distance of the two vehicles in (a).}
  \label{fig:sameMassMerger}
\end{figure}

\clearpage
\subsection*{S11. Supplementary movies}

\subsubsection*{Movie S1: Trajectory of the heavy car (200 gr.) moving on elastic membrane: Circular orbit}

\textbf{A typical circular orbit:} A video of a robotic vehicle driving on an elastic membrane with a central depression of $9.6$ cm. Instantaneous velocity and radius ($r$) are marked with red and green arrows, respectively. The heading angle is the angle between the velocity and radius.

\subsubsection*{Movie S2: Trajectory of the heavy car (200 gr.) moving on elastic membrane: Retrograde precessing orbit}

\textbf{A typical precessing orbit (retrograde):} A video of a robotic vehicle driving on an elastic membrane with a central depression of $9.6$ cm. Instantaneous velocity and radius ($r$) are marked with red and green arrows, respectively. The tracking shows that the apsis of the orbit is rotating in the opposite direction of the orbit.

\subsubsection*{Movie S3: Trajectory of the light car (45 gr.) moving on elastic membrane: Prograde precessing orbit}

\textbf{A typical precession orbit (prograde):} The lighter vehicle’s orbit undergoes a prograde precession, i.e. the vehicle and the periapsis rotate clockwise. The mass of the vehicle is about one quarter the mass of the vehicle used in Movie S1 and S2. As predicted by the theory, a radial attraction $k(r)$ decreasing with $r$ enabled by a lighter vehicle leads to the precession with the same sign of orbit, as opposed to the precession in Movie S2. 

\subsubsection*{Movie S4: Deformation-only induced motion}
In this movie, the membrane deformation is created by a human-controlled meter stick. The motion of the vehicle re-oriented by this deformation shows the deformation itself can act as a force to affect the motion of an object on the membrane.

\subsubsection*{Movie S5: Deformation-induced merger}
In the first part, both panel shows the trajectories of two vehicles moving on the membrane at the same time. The comparison is made regarding the mass ratio between the two vehicles: when the leading vehicle is heavy enough ($m_{21}=1.37$), the two vehicles eventually merge while the $m_{21}=1.00$ fails to merge.
In the second part, the video on the right panel shows the virtual superimposition of independent runs of the two vehicles with the same mass ratio as the left panel to show that the substrate-mediated interaction is indeed making the two vehicles interact.

\subsubsection*{Movie S6: Controlling speed with tilt angle to avoid collisions}
Each video shows the trajectories of the IMU-controlled vehicle (white chassis, solid line) and uncontrolled vehicle (gray chassis, dashed line) when a particular control magnitude $A=0,2,4,8$ are used.
\end{document}